\documentclass{article}

\usepackage{arxiv}

\usepackage{authblk}
\usepackage{graphicx} 
\usepackage{booktabs}
\usepackage{multirow}
\usepackage{amsmath}
\usepackage{pifont}
\usepackage{caption}
\usepackage{subcaption}
\usepackage{float}
\usepackage{placeins}
\usepackage[printonlyused, nohyperlinks]{acronym}
\usepackage{tcolorbox}
\usepackage{multicol}
\tcbuselibrary{breakable}

\usepackage{listings}
\usepackage{xcolor}
\usepackage{hyperref}

\definecolor{codegreen}{rgb}{0,0.6,0}
\definecolor{codered}{rgb}{1,0,0}
\definecolor{codegray}{rgb}{0.5,0.5,0.5}
\definecolor{codepurple}{rgb}{0.58,0,0.82}
\definecolor{backcolour}{rgb}{0.95,0.95,0.92}

\lstdefinestyle{mystyle}{
  backgroundcolor=\color{backcolour},
  commentstyle=\color{codegreen},
  keywordstyle=\color{magenta},
  numberstyle=\tiny\color{codegray},
  stringstyle=\color{codepurple},
  basicstyle=\ttfamily\footnotesize,
  breakatwhitespace=false,         
  breaklines=true,                 
  captionpos=b,                    
  keepspaces=true,                 
  numbers=left,                    
  numbersep=5pt,                  
  showspaces=false,                
  showstringspaces=false,
  showtabs=false,                  
  tabsize=2,
  escapeinside={(*@}{@*)}
}

\usepackage[numbers]{natbib}

\usepackage{array}
\newcolumntype{P}[1]{>{\centering\arraybackslash}p{#1}}

\title{Virchow2: Scaling Self-Supervised Mixed Magnification Models in Pathology}
\author[1]{Eric Zimmermann}
\author[2]{Eugene Vorontsov}
\author[2]{Julian Viret}
\author[2]{Adam Casson}
\author[2]{Michal Zelechowski}
\author[2]{George Shaikovski}
\author[1]{Neil Tenenholtz}
\author[1]{James Hall}
\author[2]{David Klimstra}
\author[2]{Razik Yousfi}
\author[2]{Thomas Fuchs}
\author[1]{Nicol\`o Fusi}
\author[2]{Siqi Liu\textsuperscript{\textdaggerdbl}}
\author[1]{Kristen Severson\textsuperscript{\textdaggerdbl}}
\affil[1]{Microsoft Research, Cambridge, MA United States}
\affil[2]{Paige, NYC, NY United States}
\date{}

\begin{document}

\maketitle

\def\thefootnote{\textdaggerdbl}\footnotetext{ These authors jointly supervised the work\\}\def\thefootnote{\arabic{footnote}}
\def\thefootnote{\textdaggerdbl}\footnotetext{ Corresponding author. siqi.liu AT paige DOT ai}\def\thefootnote{\arabic{footnote}}

\def\thefootnote{\textdaggerdbl}\footnotetext{ Corresponding author. kseverson AT microsoft DOT com}\def\thefootnote{\arabic{footnote}}

\begin{abstract}
    Foundation models are rapidly being developed for computational pathology applications. However, it remains an open question which factors are most important for downstream performance with data scale and diversity, model size, and training algorithm all playing a role. 
    In this work, we propose algorithmic modifications, tailored for pathology, and we present the result of scaling both data and model size, surpassing previous studies in both dimensions. We introduce three new models: Virchow2, a 632 million parameter vision transformer, Virchow2G, a 1.9 billion parameter vision transformer, and Virchow2G Mini, a 22 million parameter distillation of Virchow2G, each trained with 3.1 million histopathology whole slide images, with diverse tissues, originating institutions, and stains. We achieve state of the art performance on 12 tile-level tasks, as compared to the top performing competing models. Our results suggest that data diversity and domain-specific methods can outperform models that only scale in the number of parameters, but, on average, performance benefits from the combination of domain-specific methods, data scale, and model scale.\\

    \textbf{Virchow2 Model Weights}: \url{https://huggingface.co/paige-ai/Virchow2}

    \textbf{Virchow2 Azure AI Studio}: \url{https://aka.ms/virchow2}
\end{abstract}

\section{Introduction}\label{sec:introduction}

Stained tissue microscopy slides, also known as whole slide images (WSIs), are routinely collected as part of the standard of care in cancer. The digitization of these slides, at increasingly large scales, has enabled a revolution in computational pathology (CPath) towards the use of foundation models. The interest in and success of foundation models are primarily driven by the ability to learn general representations from vast amounts data that include diverse stains, tissue types, and disease without task-specific labels via self-supervised learning approaches. The use of these models represents a paradigm shift in CPath, where previously, expert systems were trained on a single tissue type with expensive, curated labels for a single task. The generalizability and robustness of a foundation model is therefore desirable for the pathology domain as there are many tasks such as diagnosis, disease subtyping, biomarker quantification, estimation of treatment response, and survival prediction. Motivated by these factors, there have been recent efforts to collect large pathology image datasets and subsequently several foundation models have been proposed ~\cite{wang2022transformer,ciga2022self,azizi2023robust,chen2024towards,vorontsov2024virchow,dippel2024rudolfv,filiot2023scaling, Xu2024}. 

Scaling up the dataset and model size has been observed to significantly boost the performance and transferability of foundation models in other domains~\cite{kaplan2020scaling, zhai2022scaling, openai_gpt-4_2023}. In the natural image domain, foundation models use millions of images (e.g. ImageNet~\cite{deng2009imagenet}, JFT-300M~\cite{sun2017revisiting} and LVD-142M~\cite{oquab2024dinov2}) to train models with hundreds of millions to billions of parameters (e.g. ViT~\cite{dosovitskiy2020image}). Early works~\cite{chen2024towards,vorontsov2024virchow} have suggested these trends exist in CPath as well, both in terms of dataset and model size, however the topic remains underexplored. Furthermore, the extent to which scaling trends observed in the natural image domain hold without adaptation to the unique aspects of CPath data is unknown.

\begin{figure}[htb]
    \centering
    \includegraphics[width=1.0\textwidth]{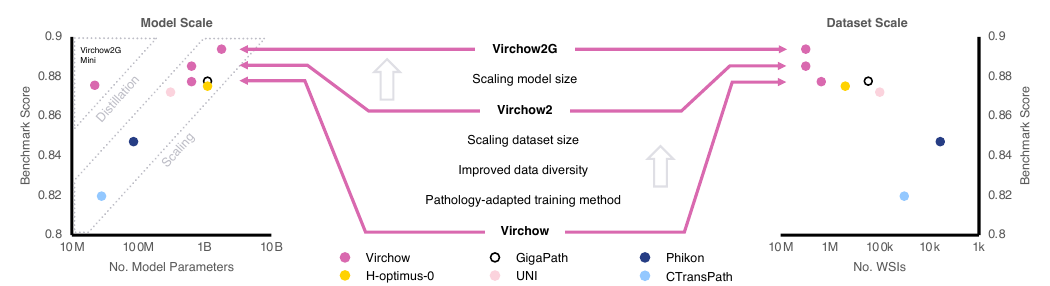}
    \caption{Virchow2 and Virchow2G adopt a pathology-adapted training method, scale up model and data size, and increase data diversity as compared to Virchow. Gains in the average weighted F1 score across 8 tile-level benchmarks using public data (see Tab.~\ref{tab:OOD-results}) are shown with respect to two scaling axes: model size (left), in terms of the number of parameters, and dataset size (right), in terms of the number of whole slide images (WSIs). Virchow2 improves over Virchow with pathology-specific modifications to the training method while scaling up the dataset from 1.5M to 3.1M WSIs, with increased diversity. Virchow2G then scales the model size from 632M to 1.9B parameters (ViT-H to ViT-G). Benchmark performance across foundation models in pathology appears to scale with model and dataset size. Distilling Virchow2G from 1.9B parameters to 22M parameters produces Virchow2G Mini (top left), an efficient model that is competitive with Virchow (632M parameters), H-optimus-0, and GigaPath (both 1.1B parameters) at a small fraction of the parameters. For all models except CTransPath, we concatenate the class token with the mean of the patch token, improving performance.}
    \label{fig:teaser}
\end{figure}

In this work, we explore scaling along both the data and model size axes, while adjusting the training method with domain-inspired modifications (Fig.~\ref{fig:teaser}). We present Virchow2 and Virchow2G, mixed magnification foundation models for computational pathology. Virchow2, a 632M parameter vision transformer (ViT-H) expands on our work developing Virchow\footnote{Virchow model weights are available for download at \url{https://huggingface.co/paige-ai/Virchow}}~\cite{vorontsov2024virchow}, scaling the dataset size from 1.5M WSIs from a single institution with routine hemotoxylin and eosin (H\&E) staining, to 3.1M WSIs from globally diverse institutions with diverse staining such as immunohistochemistry (IHC)\footnote{Virchow2 model weights are available for download at \url{https://huggingface.co/paige-ai/Virchow2}}. Virchow2G builds on Virchow2 by increasing the model size to 1.9B parameters (ViT-G).

We evaluate performance on various in-domain and out-of domain benchmarks at multiple magnifications and demonstrate the benefits of our domain-inspired adaptations at scale.
\section{Background \& related work}
\label{sec:related_work}

\subsection{Self-supervised learning}
Self-supervision is a learning paradigm  where generalizable features are learned from data using a pretext task, an objective that is constructed from implicit information contained in an input. Joint-embedding self-supervised learning (JE-SSL) methods are a subclass of SSL approaches that pose the learning objective in terms of alignment and diversity~\cite{bordes2023democratizing, wang2022understanding}. Alignment is accomplished by encouraging features, otherwise known as embeddings, of pairs or sets of samples generated from the same source image via the application of an augmentation policy to be close to one another~\cite{chen2020simple, zbontar2021barlow, bardes2022vicreg}. Diversity provides the necessary support to learn representations that avoid collapse or trivial solutions across the entire set of observations ~\cite{wang2022understanding, chen2020exploring, caron2021emerging, grill2020bootstrap, zbontar2021barlow}. These methods, in particular self-distillation variants, have emerged as powerful techniques in no-tuning transfer, both in natural images and in CPath~\cite{grill2020bootstrap, caron2021emerging, oquab2024dinov2, he2021masked, vorontsov2024virchow, kang2023benchmarking}.

\subsection{CPath foundation models}

In the past two years, at least ten CPath foundation models have been proposed in the literature (see Tab.~\ref{tab:supp_hist_FM} for an overview). The CTransPath model~\cite{wang2022transformer} was the first and trained a 28M parameter Swin transformer model using MoCoV3~\cite{chen2021empiricalstudytrainingselfsupervised} using 15M tiles from 32K WSIs. Since that time, two works~\cite{ciga2022self,azizi2023robust} have proposed models using SimCLR~\cite{chen2020simple}, but the vast majority~\cite{filiot2023scaling,chen2024towards,vorontsov2024virchow,Xu2024,nechaev2024hibou,hoptimus0,dippel2024rudolfv} have used DINOv2~\cite{oquab2024dinov2} or the related iBOT~\cite{zhou2021ibot}. The prevalence of DINOv2 appears to be largely motivated by its success with natural images as few studies have released comparative analyses for the choice of algorithm. Kang et al.~\cite{kang2023benchmarking} compared four algorithms,  DINO~\cite{caron2021emerging}, MoCo v2~\cite{chen2020improved}, SwAV~\cite{caron2020unsupervised} and Barlow Twins~\cite{zbontar2021barlow} and found no clear best method. Conversely, Chen et al.~\cite{chen2024towards} found that the choice of algorithm has a large effect on performance, observing that MoCoV3 significantly underperformed DINOv2 in CPath, despite the two approaches having comparable performance on natural image tasks.

Although CPath foundation models have largely aligned on their use of DINOv2, they are quite diverse in terms of the training dataset. Earlier works~\cite{wang2022transformer,ciga2022self,azizi2023robust,filiot2023scaling,kang2023benchmarking} relied primarily on open access data, specifically The Cancer Genome Atlas (TCGA)~\cite{weinstein2013cancer}, a repository of approximately 30K WSIs. Models were trained using this data with varying numbers of tile samples ranging from 4-50M. More recently~\cite{chen2024towards,dippel2024rudolfv,vorontsov2024virchow,hoptimus0,nechaev2024hibou,juyal2024pluto}, proprietary datasets ranging from 100K-1M WSIs have emerged, allowing for training datasest sizes to reach billions of tiles. It is worth noting that characterizing data scale in CPath is non-obvious as various works have quantified it in terms of WSIs, tiles (crops of WSIs used for training), and patients. There are further data dimensions which are also of interest such as source institution, disease status, tissue of origin, stain, and scanner.

\section{Adapting self-supervised learning to the pathology domain}\label{sec:methods}

Unlike natural images, which are a 2D projection of a 3D scene, CPath images are acquired from essentially 2D slices of tissue that are stained and scanned at various resolutions or magnifications (e.g. 5$\times$, 10$\times$, 20$\times$, 40$\times$) through a digitization process. The resulting image features vastly differ from those found in the natural image domain, as tissue patterns are repetitive, non-unique, pose-invariant, and contain meaningful but minimal color variation due to routine staining procedures, such as H\&E. CPath foundation models are trained with self-supervised learning on tissue tiles, i.e. image crops, as WSIs are gigapixel images that are costly to train with directly. Tile size is selected based on compute cost, and the quality of tile content can be highly variable. Despite these differences from the natural image domain, most CPath foundation models have directly used training algorithms developed for natural images and instead focus on the dataset as the primary differentiator. Given the crucial impact of augmentation, algorithm, and model size on foundation model performance, we present modifications to the DINOv2 recipe at scale. In the following sections, we address domain differences by adopting more domain-relevant augmentations and regularization techniques through the lens of feature alignment and diversification. 

\subsection{Domain-specific augmentations}\label{sec:morphology}

In the context of joint-embedding self-supervised learning, feature invariance and equivariance are learned through selective alignment of image pairs, also known as views, that are generated by perturbing an input with a set of random augmentations. The quality of features is dependent on the complexity of the pretext task as well as its relationship to the downstream objective. 

\begin{figure}[ht]
    \centering
    \includegraphics[width=1.0\textwidth]{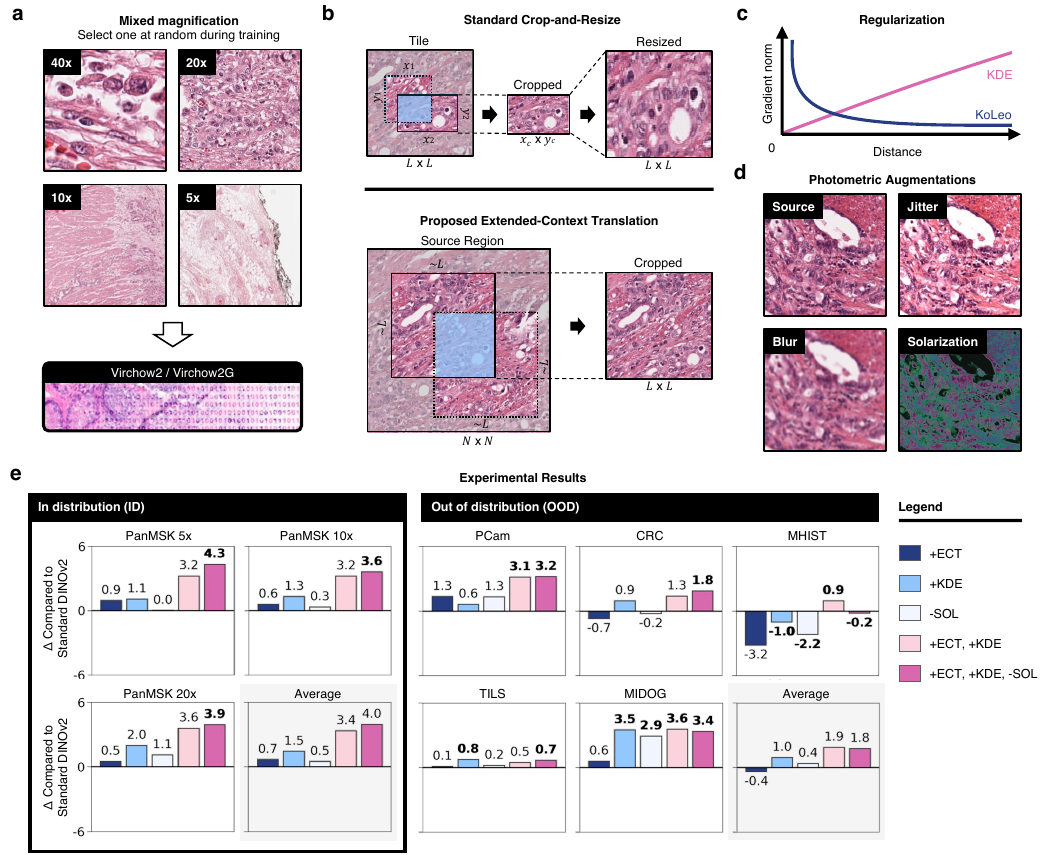}
    \caption{Highlighted modifications of DINOv2 tailored for pathology. \textbf{a.} Training samples are tiles drawn from the various available magnifications, e.g. 40x, 20x, 10x, or 5x.  \textbf{b.} Illustration of the difference between standard crop-and-resize augmentation (top) and the proposed extended-context translation (ECT) augmentation (bottom). ECT uses less resizing to avoid distorting the cell morphology. \textbf{c.} A graphic depicting the different trends of KoLeo and KDE regularizer gradient norms as a function of distance. When samples are more likely to be similar, as is the case in pathology, large gradient norms are more likely when using KoLeo. \textbf{d.} Examples of common photometric augmentations used in self-supervised learning. Solarization is hypothesized to be a poor source of color invariance for pathology applications. \textbf{e.} Results of the ablation study where each bar reports the difference between the proposal and standard DINOv2 as measured by weighted F1 score. In the majority of instances, domain-inspired changes outperform standard approaches.}
    \label{fig:methods}
\end{figure}

Augmentations are typically categorized as either photometric or geometric, and most augmentation strategies have been designed for object-centric natural images. In the majority of instances, these augmentations are directly applied in pathology applications, but there are a small number of exceptions. Examples of pathology-specific photometric augmentations include stain jitter and transfer~\cite{shen2022randstainna,gullapally2023synthetic}, where colors are augmented within and across tiles, aiming to account for differences in staining protocols and facilitate the learning of color invariances. In benchmarking studies, Tellez et al.~\cite{tellez2019quantifying}, Ciga et al.~\cite{ciga2022self}, and Gullapally et al.~\cite{gullapally2023synthetic} investigated the impact of color augmentation and demonstrated improvements in performance using domain-specific approaches. Since then, several models have employed domain-orientated approaches~\cite{wang2022transformer,kang2023benchmarking,dippel2024rudolfv}. However, the benefit of these augmentations when applied to mixed stain datasets, e.g. datasets that contain IHC, is unclear. Other photometric augmentation methods such as image solarization have been omitted in some CPath works ~\cite{pmlr-v143-faryna21a, kang2023benchmarking, dippel2024rudolfv}, as this augmentation is hypothesized to generate color profiles that are not useful for learning relevant invariances.

The random crop operation is arguably the most important geometric augmentation~\cite{moutakanni2024dontneeddataaugmentationselfsupervised}. This augmentation introduces feature co-occurrences within and across samples. These co-occurrences may be explicitly generated based on the augmentation parameters which in turn govern the amount of expected overlap ~\cite{han2022augmentation,huang2021towards} or implicitly generated based on the data distribution. The crop augmentation is paired with a resize operation due to a fixed input size and a desire for scale invariance (the resulting combination is referred to as crop-and-resize). As cell morphology plays an important role in understanding tissue structure and disease, depending on the aspect ratio of the crop, a resize may introduce unwanted distortions, affecting tissue and cell shapes.
Indeed, Ciga et al.~\cite{ciga2022self} investigated the impact of random cropping on model performance and found less random cropping generally improved performance although the largest observed improvement in any setting was 5\%. Chen et al.~\cite{chen2024towards} deviates from DINOv2's default parameters and explores less aggressive crop-and-resize scale parameters without changing aspect ratio.
As noted above, one unique aspect of pathology data is that tiles are sourced from larger WSIs. As a result of this workflow, image size is a design variable and boundaries can be extended without the need for padding or resizing. Because of this, we propose a translation-like augmentation that uses a larger source field of view to create crops with minimal resizing thereby minimizing morphological distortions while maintaining the same output size and expected overlap between image views. This operation differs from typical translations as it does not introduce imaging artifacts along boundaries, i.e. padded values. It also benefits from extending the relative field-of-view of the WSI without computational overhead. To demonstrate how this is possible, consider replacing the $L \times L$-sized training tile with an $N \times N$-sized source region and a target $L \times L$ tile to be sub-sampled within the larger window, where $N > L$ (see Fig.~\ref{fig:methods}). The particular value for $N$ can be selected based on the desired intersection over union of views. This alternative augmentation, which we refer to as extended-context translation (ECT), can serve as a drop-in replacement for traditional crop-and-resize approaches and can be combined with other augmentations such as photometric augmentations.

\subsection{Accounting for tissue redundancy}\label{sec:redundancy}

Self-supervised alignment tasks are inherently unstable and require additional regularization to encourage diversity and avoid dimension collapse. DINOv2 encourages diversity through contrastive and non-contrastive objectives. Non-contrastive diversity is achieved using asymmetric centered and sharpened distillation between the student and teacher models, while contrastive diversity aims to maximize differential entropy using KoLeo. Different diversity regularization objectives often share similar optima such as uniformity of the hypersphere; however, the dynamics of each method throughout training are considerably different and introduce practical challenges that must be addressed.

Feature diversity is dependent on the training data distribution and training methodology. In CPath, the likelihood of contrasting two similar tissue tiles may be high, unlike when contrasting images from large uncurated natural image datasets. Therefore, any contrastive diversity estimators should be constructed under the assumption that features are not independent, may be very close together, and cannot be arbitrarily separated. DINOv2 approximates differential entropy using KoLeo defined as 
\begin{equation}
    H_\textrm{KoLeo}(f) = \frac{1}{n}\sum_{i=1}^n \min_{j \neq i} \log d(\mathbf{z}_i, \mathbf{z}_j),
    \label{eqn:koleo}
\end{equation}
where $i$ indexes the samples in a batch of size $n$, $d$ is a distance measure and $\mathbf{z}$ is an embedding normalized to the hypersphere~\cite{sablayrolles2019spreading}. The inclusion of an entropy maximization objective using KoLeo aims to spread out learned embeddings via a convex regularizer. While this method has demonstrated value in natural image pretraining settings for linear and nearest neighbour classification objectives where sample diversity is high ~\cite{oquab2024dinov2}, it is clear that when two samples are very similar, and distance approaches zero, the loss approaches infinity. In practice, KoLeo is stabilized with the addition of an $\epsilon > 0$ and is implemented such that samples are not compared across devices. These factors introduce numeric and hardware dependent hyperparameter that are unexplored.

Any entropy estimator can be used as an effective replacement in order to mitigate the issues associated with KoLeo in a setting where features may be clustered, if the estimator is bounded for nearby points. We select the kernel density estimator (KDE), defined as
\begin{equation}
    H_\textrm{KDE}(f) = - \frac{1}{n} \sum_{i=1}^n  \log \sum_{j = 1}^n k(\mathbf{z}_i, \mathbf{z}_j),
    \label{eqn:kde-entropy}
\end{equation}
where \(k\) is an appropriate kernel function and all other terms are as defined above. Self-comparisons ensure the the gradients are bounded. Similar objectives have been used in various other methods~\cite{yeh2022decoupled, wang2022understanding}. The kernel density estimator has different regularization effects for local behaviors, i.e. for samples with large and small distances, but maintains the population effect of promoting diversity. Examples of kernels include Gaussian, von Mises-Fisher, inverse multiquadratic, or Laplacian kernel each of which allows for different repulsion characteristics that have bounded gradients which can also vanish for a given pair. In our work, density estimation for entropy regularization is computed using an unnormalized von Mises-Fisher (vMF) kernel, 
\begin{equation}
    k_\textrm{vMF}(\mathbf{x}, \mathbf{y}) = \exp(\kappa \mathbf{x}^\top\mathbf{y}),
\end{equation}
where $\kappa$ is a scaling constant also known as the concentration. The vMF kernel is selected because of its favorable computational qualities and its demonstrated success in encouraging diverse embeddings~\cite{wang2022understanding, Borodachov2019DiscreteEO}.

\section{Stabilizing self-supervised vision transformers at scale}\label{sec:scale}

Vision transformers are notoriously challenging to train. Architecture based instability in self-supervised learning protocols have been empirically correlated with high gradient norms occurring in the patch embedding layer at the stem of the network for large batch sizes~\cite{chen2021empiricalstudytrainingselfsupervised}. As model size is increased, large gradient norms become increasingly problematic across layers of the ViT and partial failures may not be recoverable without the manual intervention of a complete rollback prior to the event. This is costly, time consuming, and may not guarantee success. To address the architectural challenges associated with training large models, dual patch normalization (DPN)~\cite{kumar2023dualpatchnorm} and query-key normalization (QKN)~\cite{dehghani2023scalingvisiontransformers22} have been introduced in the literature and are adopted in our work. Both normalization techniques aim to limit problematic gradients, aside from gradient clipping techniques, by standardizing activations throughout the network. The inclusion of DPN circumvents the need to freeze the randomly initialized weights of the linear projection in the patch embedding layer.

Optimizing ViTs requires adaptive methods like AdamW and its variations~\cite{Balestriero2023ACO}. The stability of AdamW depends on the efficacy of the warm-up phase, on the learning rate, and on the selection of momentum parameters. Gradient spikes have been correlated with the ratio of first and second moments for an update step which provide insights on when a collapse event is likely to occur~\cite{molybog2023theoryadaminstabilitylargescale, wortsman2023stablelowprecisiontraininglargescale}. For example, instability has been observed in the weights of the patch embedding layer \cite{chen2021empiricalstudytrainingselfsupervised, wortsman2023stablelowprecisiontraininglargescale}. It is possible to reduce these issues by lowering the second moment hyperparameter but this may incur reduced downstream task performance. StableAdamW is explored as a means of stabilizing weight updates when the ratio between the first and second moments are too large (such as when nearly converged weights suddenly need a correction during a long training run) by mitigating large update steps with root-mean-square clipping \cite{wortsman2023stablelowprecisiontraininglargescale}.

Aside from architectural and optimization instability, self-supervised methods commonly suffer from partial or complete dimension collapse as a result of the imbalance between alignment and diversity objectives. Like DINO, DINOv2 primarily avoids collapse over class and patch tokens using an asymmetric centered and sharpened cross-entropy loss between the student and teacher networks. Stability over the distillation objective is highly dependent on the sharpness of the distributions which is controlled by temperature parameters. DINO studied the effects of varying teacher temperature for a fixed student temperature. Results demonstrated that higher teacher temperatures yield better transfer performance but require a warm-up period to avoid collapse, whereas a lower choice of temperature produced marginal performance drops with increased stability \cite{caron2021emerging}. Moreover, registers~\cite{darcet2024visiontransformersneedregisters} may be added mitigate the possibility of encountering high norm tokens as model size increases. High norm tokens are explored in object-centric settings where background information is redundant and of lesser importance. In contrast, pathology image tiles filtered to the tissue foreground have no such bias, thus the addition of registers is precautionary.

\section{Ablations investigating learning in the pathology domain}\label{sec:ablations}

We present an ablation study to evaluate modeling choices for the pathology domain. In order to understand the performance impacts of various augmentations and regularization methods prior to model and data scaling, a sweep is performed at a smaller scale on limited data to assess the viability of each technique. The goal of the ablation is to gain better insights into how different elements of a complex training procedure are coupled, despite the possibility that any performance gains or losses attributed to these changes may not be a factor at scale for long training horizons. The axes of exploration are in terms of geometric and photometric augmentations through the use of crop-and-resize and ECT, as well as the presence of solarization, while regularization is explored through the use of KoLeo and KDE.

\subsection{Training data}\label{sec:ablation-data}

The ablation training dataset is comprised of 1.5M H\&E stained WSI from 120K MSKCC patients scanned at $5\times, 10\times, 20\times, 40\times$ resolutions across 17 tissue types (the original Virchow training set~\cite{vorontsov2024virchow}). WSI are subdivided into $224\times224$ or $392\times392$ non-overlapping tiles and filtered to include a minimum tissue coverage of $45\%$ as determined by a hue, saturation, and value (HSV) filter with an acceptance criterion in ranges [90, 180], [8, 255], [103, 255], respectively.

\subsection{Pretraining metholodology}\label{sec:ablation-description}

A ViT-B/16 is trained using variations of DINOv2~\cite{oquab2024dinov2} on $224\times224$ global views and $96\times96$ local views sampled from image tiles. The method uses a batch size of $1024$, a stochastic depth drop rate of $0.4$, and the AdamW optimizer with a learning rate equal to $2 \times 10^{-4}$ on a cosine schedule for approximately 112K iterations, or equivalently 115M tiles sampled from the dataset detailed in section \ref{sec:ablation-data}, using 16 Nvidia V100 GPUs. The student DINO and iBOT temperature is set to $0.1$ while the teacher temperature is warmed up from $0.04$ to $0.07$ over a period of 12K iterations.

The augmentation policy is composed of horizontal and vertical flips, grayscale, and color jitter, while the crop-and-resize is interchanged with ECT. The crop-and-resize operation is applied to $224\times224$ tiles and follows the default parameters used in DINOv2, which include scale ranges of $(0.32, 1.0)$ and $(0.05, 0.32)$ for global and local views with an aspect ratio range of $(0.75, 1.33)$. ECT is applied to regions of size $392\times392$ pixels and uses a base scale range of $(0.9, 1.1)$ and aspect ratio range of $(0.95, 1.05)$. The scale range is then adjusted using the relative ratio of the source and target size for both global and local views.

Entropy regularization is performed using KoLeo, a differential method, or KDE, a resubstitution method. Density estimation is performed using the vMF kernel and the concentration is set to 5 for all experiments based on empirical results in the literature.

\subsection{Tile benchmarks}\label{sec:eval-tasks}
A mix of seven in-distribution (ID) and out-of-distribution (OOD) tile level classification tasks across various magnifications are used to evaluate the quality and separability of the learned features of the teacher network via a linear probe protocol.

\textbf{PanMSK} is an in-distribution cancer detection task performed at 5$\times$, 10$\times$, and 20$\times$ magnifications. 1,196,171 $224 \times 224$ pixel tile samples, sourced from 3,999 WSI across 2534 patients from MSKCC representing 17 tissue types, are labeled either cancer or benign \cite{vorontsov2024virchow}. To ensure a sufficient sample size for evaluation, no tissue (except for endometrial) has fewer than 44 WSIs with cancer from no fewer than 25 patients (max 139, median 43).

\textbf{PCam} is a public (PatchCamelyon) dataset of 327,680 lymph node images labeled as cancer or benign \cite{bejnordi2017diagnostic,veeling2018rotation}. Images are upsampled from $96 \times 96$ at $10\times$ magnification to $224 \times 224$ for analysis.

\textbf{MHIST} is a public dataset of 3152 colorectal polyp images ($5\times$ magnification $224 \times 224$ pixels) labeled benign or precursor \cite{wei2021petri}, used as a ``minimalist histopathology image analysis dataset'' (MHIST). Since a validation data split is not provided, we randomly split 10\% of the training set into a validation set.

\textbf{CRC} is a public dataset with colorectal cancer (CRC), composed of 100,000 colorectal images ($20\times$ magnification $224 \times 224$ pixels) classified into 9 morphologies~\cite{kather_jakob_nikolas_2018_1214456}.

\textbf{MIDOG} is a binary mitotic cell detection task that we created from the public MItosis DOmain Generalization (MIDOG++) dataset, containing 21,806 mitotic and non-mitotic events labeled on 503 $7K \times 5K$ pixel WSI regions from several tumor, species, and scanner types at $40\times$ magnification~\cite{aubreville2023comprehensive, vorontsov2024virchow}. This data was converted into a binary classification task by expanding each $50 \times 50$ pixel annotation to $224 \times 224$ regions and then randomly shifting in the horizontal and vertical regions such that the event is not centered in the tile. All negative instances that overlapped with positive instances were removed from the dataset. 

\subsection{Linear evaluation protocol}

Performance is assessed by training a linear classifier on z-score normalized frozen embeddings generated from non-augmented samples. The classifier is trained with a batch size of 4096 using stochastic gradient descent with a cosine learning rate schedule from $1\times10^{-2}$ to $0$ for a total of 12.5K iterations, to maximize convergence. The classifier checkpoint achieving the lowest validation loss was selected for evaluation on a holdout test set. All benchmark datasets are partitioned to include a validation set if not provided by the public resource. If no validation set exists, a random split of $10\%$ is created from the training set.

\subsection{Ablation results}
Fig.~\ref{fig:methods}e  presents the results of the ablation analysis (see Tab.~\ref{app-tab:id}-\ref{tab:abalation_ood_cls_only} in Appendix~\ref{sec:appendix_ablation} for additional details). In a majority of tasks (5/7), using all three domain-inspired changes, results in the best performance as well as the best average performance on in-distribution tasks. Using only ECT and KDE regularization has the best performance of 3/7 tasks and the highest average performance on out-of-distribution tasks, although the gap with all three is only 0.002. Interestingly, we observe that using only ECT leads to a small drop in performance while using only KDE regularization leads to a small improvement in performance on out-of-distribution tasks. Combining the two leads to a pronounced improvement, as noted above. All changes improved in-distribution tasks.

\section{Results of large-scale training}

We improve on the training recipe of Virchow with pathology specific tuning and scale up the training first along the data axis (Virchow2) and then along the model axis (Virchow2G). Afterwards, we explore the effectiveness of scaling up in order to scale down by using Virchow2G as a teacher to train a significantly smaller yet high-quality student (Virchow2G Mini).  We evaluate these models against other foundation models in pathology on tile-level benchmark tasks.

\subsection{Training data}\label{sec:data}

Virchow2 is trained on a dataset of 3.1M biopsy and resection WSIs processed from 225,401 patients across $5\times$, $10\times$, $20\times$, $40\times$ magnifications, with both hematxylin and eosin (H\&E) and immunohistochemical (IHC) staining. This dataset extends the 1.5M WSI dataset that Virchow was trained on and is no longer restricted to 17 tissue types (indeed, it contains nearly 200 recorded tissue types). In addition to samples collected by the Memorial Sloan Kettering Cancer Center (MSKCC) 15\% of the WSIs and 57\% of the patients are sourced from diverse institutions worldwide that submitted challenging cases to MSKCC for consultation. 

\begin{figure}[ht]
    \centering
    \includegraphics[width=1.0\textwidth]{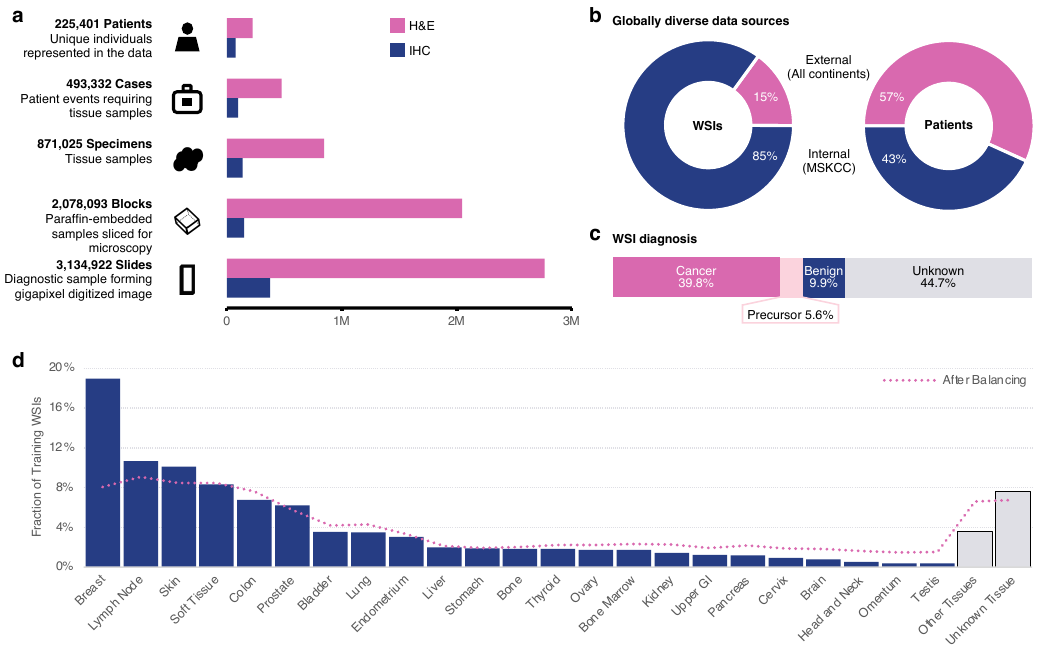}
    \caption{Overview of the training data. \textbf{a.} A breakdown of the scale of the training dataset with 3.1M whole slide images (WSIs), showing the distribution of slides with hemotoxylin and eosin (H\&E) and immunohistochemical (IHC) stains. \textbf{b.} The data are globally diverse, with 15\% of the slides and 57\% of the patients sourced from many institutions across all continents. These slides were submitted to MSKCC for review. \textbf{c.} The distribution of most severe diagnoses of the WSIs, in decreasing order of severity: cancer, precursor, benign, unknown. \textbf{d.} The top identified tissue types shown as a fraction of the WSIs. There are over 150 additional tissue types in the long tail, summarized here as `Other Tissues' (shown gray, along with `Unknown Tissue', on the right). Online data balancing when training Virchow2G affected these tissue distributions as shown by the magenta dotted line.}
    \label{fig:data}
\end{figure}

Non-overlapping region tiles of size $392\times392$ pixels with at least 65\% tissue by area are used for model training. Note that Virchow2 and Virchow2G extract $224\times224$ crops from these tiles. Tissue segmentation is performed with a trained fully-convolutional network, with post processing via Otsu thresholding with  thresholds of $(0.4, 0.5)$.

Due to the long-tail nature of the data, balancing is explored in terms of tissue type, diagnosis, stain, and magnification. Metadata is not complete for all samples and the presence of cancer, precursor, and neoplastic tissue does not imply the lack of benign tiles within these large samples. Given these sources of noise, we aim to balance with the best available data. Cancer is sampled at 40\% and precursor is sampled at 15\%, which closely matches the observed frequency of precursor samples in the dataset (17.9\%, where 12.3\% contain both cancer and precursor and 5.6\% contain only precursor). Benign and benign non-neoplasms are sampled at approximately half their observed frequencies for a total of 8\%. Benign neoplasms are sampled at 2\%. Samples with unknown diagnosis are sampled at 35\% (see Fig.~\ref{fig:methods}c for the observed distribution). Tissue type balancing is performed to flatten out the distribution while maintaining the proportion of IHC in an H\&E dominant dataset. This approach primarily decreases the prevalence of breast and increases the prevalence of tissues with small sample size and unknown tissue (see Fig.~\ref{fig:data}d). Magnification balancing is performed to increase uniformity in batches during training under the assumption that each resolution may contain unique feature sets that should be contrasted against one another. 40\% of WSI do not have 40$\times$ magnifications, therefore these slides are sample 1.5$\times$ more often than those which do not. Sampling of the magnifications is also balanced to yield a final approximate distribution of 20\%, 40\%, 20\%, and 20\% for 40$\times$, 20$\times$, 10$\times$, 5$\times$, respectively.

\subsection{Pretraining methodology}\label{sec:methodology}

Two models are trained for a full schedule using insights from the ablations detailed in section \ref{sec:ablation-description}. Both models include ECT instead of crop-and-resize augmentations as well as KDE rather than KoLeo for diversity regularization. Global tiles of size $224\times224$ pixels and local tiles of size $98\times98$ are extracted from the $392\times392$ extended context source regions, following the DINOv2 multi-view recipe. Each model is trained on $512$ Nvidia V100 GPUs. Further configuration details are found in appendix \ref{tab:v2-hyperparameters}.

Virchow2, a ViT-H/14 with 4 registers, explores the effects of increased data scale and diversity across a mix of magnifications paired with the algorithmic changes proposed. Virchow2 is trained with solarization on a cosine schedule and is optimized with AdamW for a base learning rate equal to $2\times10^{-4}$ with a batch size of $4096$ over a total of 2B unbalanced tiles. 

Following the modifications made to the ViT architecture to maximize compute efficiency~\cite{oquab2024dinov2}, we introduce Virchow2G, a ViT-G/14 that adopts similar architectural changes. Virchow2G increases the embedding dimension from 1,664 to 1,792 to be a multiple of 256 and increases the number of attention heads from 16 to 28 in order to have 64 dimensions per head. Additionally, Virchow2G adopts additional normalization via DPN and QKN and has 8 registers. Virchow2G is trained without solarization on a cosine schedule and is optimized with StableAdamW for a base learning rate equal to $1\times10^{-4}$ and a batch size of $3072$ over a total of 2 billion balanced tiles. For stability, the second moment optimizer hyperparameter was reduced from $0.999$ to $0.95$. Additionally, both iBOT and DINO head teacher temperatures are reduced from $0.07$ to $0.04$. Further configuration details are found in appendix \ref{tab:v2g-hyperparameters}.

\subsection{Distillation methodology}
We distilled Virchow2G Mini, a ViT-S/14, from Virchow2G following the DINOv2 distillation methodology~\cite{oquab2024dinov2}.
The ViT-S/14 is the standard variation of the architecture but with 4 registers and without the other modifications used in Virchow2G in order to focus on maximizing inference efficiency (i.e. GELU MLP instead of SwiGLU, no DPN, no QKN). Virchow2G Mini is trained with StableAdamW for a base learning of $2\times10^{-4}$ with a batch size of $2048$ over a total of 1B balanced tiles. The full configuration details can be found in appendix \ref{tab:v2gmini-hyperparameters}.

\subsection{Embedding generation}
For a 224 × 224 input tile image, we choose one of two configurations of embeddings: (1) the class token (CLS-Only) and (2) the class token concatenated with the mean across all 256 of the patch tokens (CLS+Mean). The former is the recommended configuration of H-optimus-0~\cite{hoptimus0}, Prov-GigaPath~\cite{Xu2024}, UNI~\cite{chen2024towards}, and Phikon~\cite{filiot2023scaling}; the latter is the recommended configuration for the Virchow models. We evaluate both CLS-Only and CLS+Mean embeddings for all models except CTransPath~\cite{wang2022transformer}, for which the mean of all tokens is used as there is no class token. The CLS+Mean configurations produces embeddings of size 2,560 (1,280 $\times$ 2) for Virchow and Virchow2, 3,584 (1,792 $\times$ 2) for Virchow2G, and 768 (384 $\times$ 2) for Virchow2G Mini.

\subsection{Tile benchmarks}
The benchmarking protocol is the same as described in Sec.~\ref{sec:eval-tasks}. In addition to the aforementioned tasks, we add an additional four out-of-distribution tasks.

\textbf{CRC No-Norm} is a variation of the CRC task where the testing set does not use Macenko stain normalization. This simulates a distribution shift between the training or validation sets and the testing set.

\textbf{WILDS} is a public dataset (Camelyon17-WILDS) of 455,954 images, each at a resolution of 96 $\times$ 96 pixels at 10$\times$ magnification upsampled to 224 $\times$ 224 derived from the larger Camelyon17 dataset~\cite{koh2021wilds}. The images are labeled to indicate the presence or absence of tumor. The validation and testing data are from institutions not represented in the training data.

\textbf{DLBCL} is a binary 5-year survival prediction task from the public DLBCL-Morph dataset containing 42 tissue microarrays from 209 Diffuse Large B-Cell Lymphoma (DLBCL) cases/patients, with H\&E staining and five IHC stains (CD10, BCL6, MUM1, BCL2, and MYC.) at 40$\times$ magnification~\cite{vrabac2021dlbcl}. Due to damaged or missing cores, some stains are missing for some patients. Training, validation, and testing sets are created on the patient level.

\textbf{TCGA-TILS} is a public dataset of 304,097 images, each at a resolution of 100 $\times$ 100 pixels at 20$\times$ magnification. The images are labeled to indicate the presence of tumor-infiltrating lymphocytes (TILs) if at least two TILs are present~\cite{kaczmarzyk10dataset,abousamra2022deep,saltz2018spatial}.

\textbf{HEST-Benchmark} is a public dataset of 74 spatial transcriptomics profiles from 48 patients grouped into 10 tasks based on organ~\cite{jaume2024hest}. The aim of the task is to predict the expression levels of the 50 most variable genes from 112 $\times 112 \mu$m H\&E stained image patches, each centered on a spatial transcriptomics spot. Here we present the results of a random forest classifier following the protocol in~\cite{hest_github}, as opposed to the linear protocol described in Sec.~\ref{sec:eval-tasks}.

\subsection{Scaling results}\label{sec:results}
Virchow2 and Virchow2G, along with Virchow and other baseline foundation models, were evaluated on in-distribution (ID) tile-level benchmarks (Tab.~\ref{tab:ID-results}) and out-of-distribution (OOD) tile-level benchmarks (Tab.~\ref{tab:OOD-results}). The pathology-specific modifications to the training recipe of Virchow2, along with increased training data scale and diversity, yielded a substantial boost in ID performance, raising the average weighted F1 score from 0.944 (Virchow) to 0.966 (Virchow2). Gains in ID performance from further scaling the model size with Virchow2G were more modest, further raising the score to 0.971. OOD average weighted F1 score increased from 0.877 with Virchow to 0.885 with Virchow2 and 0.894 with Virchow2G, demonstrating the benefits of model and data scale, as well as algorithmic improvements. Indeed, Fig.~\ref{fig:teaser} demonstrates a log-linear trend in average OOD performance with model and data scale, across all models.

Although H-optimus-0 and Prov-GigaPath both use a ViT-g with 1.1B parameters, Virchow with ViT-H and 632M parameters matches or exceeds their performance on average, being surpassed only on PanMSK $20\times$, MHIST, PCam, and WILDS by H-optimus-0 and on PanMSK $5\times$, PCam, and WILDS by Prov-GigaPath. Virchow2, also with 632M parameters but with a larger training set and improved training recipe, outperforms H-optimus-0 and Prov-Gigapath on all ID (PanMSK) tasks and MHIST, and bridges the gap on WILDS. Scaling up the model size to ViT-G (1.9B parameters) with Virchow2G, results in top performance on all benchmark tasks. We note that while Virchow2G achieves a weighted F1 score of 0.948 on TILS, as compared to the top score of 0.949 (Virchow, H-optimus-0), these scores are similar and performance on this task appears to have saturated. A major source of improvement lies in the addition of mixed magnifications. Virchow2 and Virchow2G both significantly improve on ID PanMSK tasks at $10\times$ and $5\times$ magnification. High resolution $40\times$ OOD tasks such as MIDOG and DLBCL further illustrate the benefits of training on magnifications other than $20\times$.

Tab.~\ref{tab:hest_st} presents the Pearson correlation coefficient results for the HEST-Benchmark. Here we observe less consistency in the performance of the models, possibly because the official training protocol uses fixed hyperparameters for all models' embeddings, gene expression signal may not always be evident in H\&E, and low sample counts for some tasks may add noise to the evaluation. However, the average performance across all ten gene expression prediction tasks scales with model size, as shown in Fig.~\ref{fig:all_scaling}c (Appendix~\ref{sec:appendix_scaling}). Virchow2G (1.9B parameters) has the best average performance at 0.350 (best on 1/10 tasks, second best on 3/10 tasks), followed by H-optimus-0 (1.1B parameters) at 0.347 (best on 3/10 tasks, second best on 1/10 tasks). Results for all models are slightly higher in our evaluation than in the benchmark literature~\cite{jaume2024hest}, possibly due to differences in supporting software library versions or non-determinism in the evaluation code.

We use the concatenation of the class [CLS] token with the mean of the patch tokens (CLS+Mean) as the embeddings for all Virchow models. While using the class token only (CLS) is a suggested as the default for H-optimus-0, Prov-GigaPath, UNI, and Phikon, we find that CLS+Mean embeddings produce the same or better benchmark results, especially for mitosis detection (MIDOG), a localized detection task (see Appendix~\ref{app_sec:patch-tokens} for additional experimental results). Results for all tasks are shown for both embeddings configurations in Tab.~\ref{tab:ID-results}, Tab.~\ref{tab:OOD-results}, Tab.~\ref{tab:hest_st}, and Fig.~\ref{fig:all_scaling} (Appendix~\ref{sec:appendix_scaling}). Interestingly, Virchow2 performance is similar when using CLS or CLS+Mean embeddings, suggesting that CLS embeddings may be used with to the reduce storage cost; however, further task-specific validation is necessary to confirm that performance is retained in this configuration.

Our results show that benchmark performance scales with the size of the pre-trained model across diverse benchmarks (see Fig.~\ref{fig:all_scaling} in Appendix~\ref{sec:appendix_scaling}). Although large models are computationally expensive and their large embeddings are storage-intensive, they also offer a good starting point for model distillation. Previous work has shown DINOv2 distillation is parameter-efficient~\cite{oquab2024dinov2}. Indeed, using the same method to distill Virchow2G from a 1.9B parameter ViT-G to a 22M parameter ViT-S (roughly 1\% of the parameters) we obtain a model (Virchow2G Mini) that outperforms H-optimus-0 (1.1B parameters), GigaPath (1.1B parameters), and Virchow (632M parameters) in-distribution and nearly ties Virchow out-of-distribution while outperforming H-optimus-0 (Tab.~\ref{tab:ID-results}, Tab.~\ref{tab:OOD-results}).

\begin{table}[]
\centering
{ 
\begin{tabular}{llcccc} \toprule
 &  Model             & PanMSK 20x & PanMSK 10x & PanMSK 5x & Average \\ \hline \hline
 \multirow{7}{*}{\begin{tabular}{@{}c@{}}\rotatebox[origin=c]{90}{CLS+Mean}\end{tabular}}
 & Virchow2G    & \textbf{96.6}       & \textbf{97.1}       & \textbf{97.5}      & 97.0       \\
 & Virchow2     & 96.4       & 96.6       & 96.7      & 96.6       \\
 & Virchow2G Mini & 95.0 & 95.7 & 96.0 & 95.6 \\
 & Virchow     & 95.0       & 94.8       & 93.3      & 94.4       \\
 & H-optimus-0     & 95.7       & 95.3       & 93.8      & 95.0       \\
 & Prov-GigaPath & 94.2       & 94.8       & 94.8      & 94.6       \\
 & UNI           & 94.4       & 94.8       & 93.8      & 94.3       \\
 & Phikon        & 92.4       & 92.6       & 90.8      & 91.9       \\\hline
 $\dag$ & CTransPath    & 89.7       & 92.7       & 93.3      & 91.9       \\\hline
 \multirow{8}{*}{\begin{tabular}{@{}c@{}}\rotatebox[origin=c]{90}{CLS-Only}\end{tabular}}
 & Virchow2G    & \textbf{96.6}       & \textbf{97.2}       & \textbf{97.5}      & 97.1       \\
 & Virchow2     & 96.4       & 96.7       & 96.8      & 96.6       \\
 & Virchow2G Mini &  95.1&  96.1&  96.3&  95.8\\
 & Virchow     & 95.0       & 94.9       & 92.4      & 94.1       \\
 & H-optimus-0     & 95.2       & 94.7       & 93.4      & 94.4       \\
 & Prov-GigaPath & 94.0       & 94.7       & 94.4      & 94.4       \\
 & UNI           & 94.3       & 95.0       & 93.6      & 94.3       \\
 & Phikon        & 92.3       & 92.6       & 90.1      & 91.7       \\
 
 & NatImg        & 88.3       & 90.0       & 91.8      & 90.0      \\
 \bottomrule
\end{tabular} }
\caption{Weighted F1 score results for the in-distribution tile-level benchmark tasks. Best results, and predictions which are not statistically significantly different at $\alpha=0.05$ as determined by McNemar's test, are bolded for each task. Concatenating the mean of the patch tokens to the class token (CLS+Mean) tends to produce better results than when using only the class token (CLS-Only). $\dag$ CTransPath lacks a CLS token and uses the mean of all tokens. Note that results have been multiplied by 100 for improved readability.}
\label{tab:ID-results}
\end{table}

\begin{table}[]
\centering
{
\begin{tabular}{llccccccccc} \toprule
& Model            & PCam          & CRC           & {\begin{tabular}[c]{@{}c@{}}CRC\\ No-Norm\end{tabular}}    & WILDS         & TILS          & MHIST         & DLBCL         & MIDOG         & Average \\
\hline\hline
\multirow{7}{*}{\begin{tabular}{@{}c@{}}\rotatebox[origin=c]{90}{CLS+Mean}\end{tabular}}
& Virchow2G    & \textbf{94.7} & \textbf{97.3} & \textbf{97.0} & \textbf{98.8} & \textbf{94.8} & \textbf{86.4} & \textbf{62.9} & \textbf{83.6} & 89.4       \\
& Virchow2     & 93.5          & \textbf{97.4} & \textbf{96.9} & 98.7          & \textbf{94.8} & \textbf{85.9} & 60.6          & 80.4          & 88.5       \\
& Virchow2G Mini & 93.5 & 97.0 & 96.0 & 98.4 & 94.1 & 83.2 & 58.9 & 79.6 & 87.6 \\
& Virchow     & 93.3          & \textbf{97.3} & \textbf{96.8} & 97.1          & \textbf{94.9} & 83.6          & 60.2          & 78.7          & 87.7       \\
& H-optimus-0     & 93.7          & 96.9          & 92.6          & 98.6          & \textbf{94.9} & \textbf{84.6} & 59.1          & 79.6          & 87.5       \\
& Prov-GigaPath & 94.1          & 96.9          & \textbf{96.9} & 97.9          & 94.2          & 82.3          & 60.3          & 79.4          & 87.8       \\
& UNI           & 93.5          & 96.7          & 95.4          & 98.3          & 94.6          & \textbf{84.8} & 58.4          & 75.9          & 87.2       \\
& Phikon        & 91.5          & 96.3          & 88.7          & 97.0          & 94.7          & 80.9          & 55.6          & 73.0          & 84.7       \\ 
\hline
$\dag$ & CTransPath   & 87.1          & 96.2          & 84.4          & 94.7          & 93.1          & 81.6          & 54.2          & 64.3          & 82.0       \\
\hline
\multirow{8}{*}{\begin{tabular}{@{}c@{}}\rotatebox[origin=c]{90}{CLS-Only}\end{tabular}}
& Virchow2G    & \textbf{94.4} & 97.4          & \textbf{97.0} & 98.3          & 94.7          & \textbf{85.2} & \textbf{62.4} & \textbf{80.5} & 88.7       \\
& Virchow2     & 93.5          & \textbf{97.6} & \textbf{97.1} & 98.5          & 94.7          & \textbf{86.0} & 61.5          & \textbf{80.0} & 88.6       \\
& Virchow2G Mini &  93.4&  97.0&  96.2&  98.2&  93.9&  82.1&  59.9&  76.2&  87.1\\
& Virchow     & 93.4          & 97.0            & 93.2          & 96.6          & \textbf{94.8} & \textbf{83.6} & 59.1          & 76.0          & 86.7       \\
& H-optimus-0     & 93.7          & 96.3          & 94.3          & \textbf{98.6} & \textbf{94.9} & \textbf{84.8} & 58.2          & 78.1          & 87.4       \\
& Prov-GigaPath & 94.1          & 96.7          & \textbf{96.8} & 97.6          & 94.1          & 82.2          & 57.3          & 78.2          & 87.1       \\
& UNI           & 93.4          & 96.3          & 94.3          & 98.3          & 94.5          & \textbf{84.5} & 56.5          & 74.8          & 86.6       \\
& Phikon        & 90.5          & 95.9          & 88.8          & 97.2          & 94.4          & 79.6          & 54.4          & 69.9          & 83.8       \\
& NatImg        & 88.7          & 95.2          & 92.8          & 93.4          & 93.0          & 83.2          & 53.9          & 68.9          & 83.6       \\ 
\bottomrule
\end{tabular} }
\caption{Weighted F1 score results for the out-of-distribution tile-level benchmark tasks. Best results, and predictions which are not statistically significantly different at $\alpha=0.05$ as determined by McNemar's test, are bolded for each task. Concatenating the mean of the patch tokens to the class token (CLS+Mean) tends to produce better results than when using only the class token (CLS-Only). $\dag$ CTransPath lacks a CLS token and uses the mean of all tokens. Note that results have been multiplied by 100 for improved readability.}
\label{tab:OOD-results}
\end{table}

\begin{table}[]
\centering
\small
{\renewcommand{\arraystretch}{1.15} 
\begin{tabular}{llccccccccccc} \toprule
\multirow{8}{*}{\begin{tabular}{@{}c@{}}\rotatebox[origin=c]{90}{CLS+Mean}\end{tabular}}
 & Model          & IDC & PRAD & PAAD & SKCM & COAD & READ & CCRCC & HCC & LUAD & LIDC & Avg. \\ \hline \hline
 & Virchow2G                     & \textbf{0.559}                   & 0.385                    & 0.458                    & 0.632                    & 0.139                    & \textbf{0.175}                    & 0.222                     & 0.062                   & 0.588                    & 0.274                          & 0.349   \\
 & Virchow2                      & 0.539                   & 0.382                    & 0.425                    & 0.617                    & 0.127                    & 0.168                    & \textbf{0.226}                     & 0.056                   & 0.586                    & 0.273                          & 0.340   \\
 & Virchow                      & 0.545                   & 0.372                    & 0.465                    & 0.624                    & \textbf{0.159}                    & 0.133                    & 0.211                     & \textbf{0.065}                   & \textbf{0.601}                    & 0.269                          & 0.344   \\
 & H-optimus-0 & 0.555                   & 0.388                    & \textbf{0.469}                    & \textbf{0.650}                    & 0.158                    & 0.161                    & 0.222                     & 0.062                   & 0.596                    & \textbf{0.278}                          & 0.354   \\
  & Prov-GigaPath                 & 0.527                   & \textbf{0.392}                    & 0.453                    & 0.623                    & 0.147                    & 0.136                    & 0.189                     & 0.051                   & 0.574                    & 0.267                          & 0.336   \\
 & UNI                           & 0.531                   & 0.391                    & 0.444                    & 0.630                    & 0.149                    & 0.131                    & 0.183                     & 0.046                   & 0.571                    & 0.265                          & 0.334   \\
 & Phikon                        & 0.494                   & 0.388                    & 0.433                    & 0.595                    & 0.151                    & 0.123                    & 0.178                     & 0.062                   & 0.577                    & 0.259                          & 0.326   \\ \hline
 $\dag$  & CTransPath                    & 0.476                   & 0.363                    & 0.416                    & 0.557                    & 0.118                    & 0.093                    & 0.177                     & 0.063                   & 0.560                    & 0.255                          & 0.308   \\ \hline
\multirow{8}{*}{\begin{tabular}{@{}c@{}}\rotatebox[origin=c]{90}{CLS-Only}\end{tabular}}
 & Virchow2G                     & 0.547                   & 0.375                    & 0.420                    & 0.638                    & 0.140                    & 0.146                    & 0.213                     & 0.056                   & 0.590                    & 0.271                          & 0.339   \\
 & Virchow2                      & \textbf{0.563}                   & 0.379                    & 0.402                    & 0.631                    & 0.143                    & \textbf{0.169}                    & \textbf{0.224}                     & 0.060                   & \textbf{0.591}                    & \textbf{0.275}                          & 0.344   \\
 & Virchow                      & 0.529                   & 0.364                    & 0.431                    & 0.612                    & 0.150                    & 0.120                    & 0.205                     & 0.065                   & 0.584                    & 0.274                          & 0.334   \\
  & H-optimus-0 & 0.536                   & 0.378                    & \textbf{0.446}                    & \textbf{0.649}                    & \textbf{0.159}                    & 0.160                    & 0.210                     & \textbf{0.068}                   & \textbf{0.591}                    & 0.271                          & 0.347   \\
 & Prov-GigaPath                 & 0.514                   & \textbf{0.386}                    & 0.437                    & 0.577                    & 0.144                    & 0.156                    & 0.188                     & 0.057                   & 0.568                    & 0.274                          & 0.330   \\
 & UNI                           & 0.498                   & 0.371                    & 0.437                    & 0.637                    & 0.142                    & 0.160                    & 0.178                     & 0.052                   & 0.565                    & 0.261                          & 0.330   \\
 & Phikon                        & 0.468                   & 0.383                    & 0.397                    & 0.581                    & 0.136                    & 0.138                    & 0.175                     & 0.043                   & 0.566                    & 0.264                          & 0.315   \\ \hline
\end{tabular} }
\caption{Pearson correlation results for the ten HEST benchmark tasks, learned with a random forest classifier. Bold indicates the best performance and ``Avg.'' is the average. $\dag$ CTransPath lacks a CLS token and uses the mean of all tokens. Note that results have been multiplied by 100 for improved readability.}
\label{tab:hest_st}
\end{table}

\section{Discussion}

Our results suggest that foundation model performance in computational pathology continues to benefit from model and data scale. Nevertheless, substantial gains can still be achieved by methodological improvements even at small scales. We therefore built Virchow2 and Virchow2G, which scale the dataset size and then the model size, respectively, on top of modifications of the DINOv2 training setup, tailored to the pathology domain.

Our ablation results demonstrate that such modifications can significantly improve benchmark performance. Changing the KoLeo regularizer to a kernel density estimator (KDE) also boosted performance. This is likely because the conditions that produce instability with KoLeo are particularly common in pathology images: that is, tiles within a minibatch are much more likely to be similar to each other compared to ImageNet~\cite{deng2009imagenet} images within a minibatch. Furthermore, the proposed extended-context translation (ECT) is more fitting for pathology data than crop-and-resize because it avoids distorting the cell morphology which is crucial for interpreting pathology images. Although decreasing random resizing during training may reduce the model's scale invariance, we can retain it for pathology by training across multiple magnifications. Similarly, removing solarization boosted performance (especially on PCam), as also previously observed~\cite{kang2023benchmarking, dippel2024rudolfv}. Indeed, it has been shown that the effects of augmentation are often data dependent and vary across classes \cite{balestriero2022effects}, however, with a sufficiently large and diverse training dataset, the importance of photometric data augmentations diminishes~\cite{moutakanni2024dontneeddataaugmentationselfsupervised}. Nevertheless, suitable image transforms, like cropping, are still necessary between the two views of an input in joint embedding models like DINOv2. Finally, we find that different tasks benefit from different embedding configurations. High level understanding of a tissue tile can be achieved with a single ``class'' token that summarizes the tile; however, we find that some tasks (such as detecting cellular mitosis events) require localized information that may not be complete in this class representation. We observe that using the patch tokens in addition to the class token lead to significant performance boosts.

Further adjustments to the training recipe were required to allow scaling Virchow both along the data size and the model size axes. Without KDE, DINOv2 training collapsed before enough of the 3.1M slide dataset could be processed. Additional stabilizations were required to then scale up the model from ViT-H (632M parameters) to ViT-G (1.9B parameters). We note that largest models previously trained with DINOv2 had 1.1B parameters (ViT-g)~\cite{oquab2024dinov2, hoptimus0, Xu2024}. The types of challenges encountered are classified into numerical issues and collapse events. Numerical issues are determined when a training run fails due to NaNs that could not be caught and mitigated and may be due to the use of FP16 rather than BF16. These are addressed by DPN, QKN, and gradient and update clipping in StableAdamW. Collapse events are categorized by sudden spikes in a loss component when features collapse to a lower dimensional subspace that is correlated with the degradation in downstream performance. Collapse primarily occurs in the iBOT head and is mitigated by reducing the learning rate, reducing the $\beta2$ momentum parameter in StableAdamW from $0.999$ to $0.95$, and reducing the teacher temperature from $0.07$ to $0.04$ based on the guidance in DINO~\cite{caron2021emerging}. No collapse was observed in the DINO head. It is possible that diversity regularization provided sufficient support to avoid catastrophic events where asymmetric centering and sharpening was insufficient. These modifications reduce the convergence rate but allow training longer. 

Based on the scaling plots (Fig.~\ref{fig:teaser}) and performance of Virchow2 and Virchow2G, benchmark performance continues to improve with model and data scale; however, additional benchmarks on which performance has not yet saturated may be necessary to gauge further gains (like MIDOG, DLBCL, or PanMSK, all of which contain magnifications other than 20$\times$). No substantial improvements with model scale were observed on WILDS, CRC, CRC-No-Norm, or TILS beyond ViT-H (632M parameters). Indeed, F1 scores for these tasks, as well as PCam and PanMSK are between 0.95 and 0.99 (where 1.00 is the maximum), suggesting that further performance gains may be insignificant or unmeasurable due to label noise. MHIST results are difficult to interpret due to their high variance. Although PanMSK~\cite{vorontsov2024virchow} covers 17 tissue types (all the tissue types that Virchow was trained on), it does not evaluate the long tail of tissue types learned by Virchow2 and Virchow2G and it could be made more challenging by enriching it for rare cancers. Indeed, in order to quantify the benefits and impact of model and data scale, it is imperative to prepare diverse benchmarks across different tasks, as some benchmarks may saturate more quickly than others. Without a reliable suite of downstream tasks, confident claims cannot be made about the quality of a model.

Ultimately, there is still room to improve benchmark performance along the axes of model scale, data scale, and algorithmic improvements. We demonstrate the value of pathology-specific improvements, as well as training on mixed magnifications, allowing improved performance on diverse benchmark tasks towards a unified image-based foundation model.

\section{Acknowledgements}
We gratefully thank Philip Rosenfield from Microsoft and Djamilia Dierov from Paige for their contributions in making this collaboration possible, Philippe Mathieu for distributed inference support, Mark Fleishman for data support, and Wayne Hendricks and Alexander van Eck at Paige and Jim Jernigan, Lifeng Li, Ben Huntley, Wyman Chong, Oleg Losinets, and the rest of the team at MSR for infrastructure support.

\bibliographystyle{plainnat}
\bibliography{references} 

\begin{thebibliography}{64}
\providecommand{\natexlab}[1]{#1}
\providecommand{\url}[1]{\texttt{#1}}
\expandafter\ifx\csname urlstyle\endcsname\relax
  \providecommand{\doi}[1]{doi: #1}\else
  \providecommand{\doi}{doi: \begingroup \urlstyle{rm}\Url}\fi

\bibitem[Abousamra et~al.(2022)Abousamra, Gupta, Hou, Batiste, Zhao, Shankar, Rao, Chen, Samaras, Kurc, et~al.]{abousamra2022deep}
Shahira Abousamra, Rajarsi Gupta, Le~Hou, Rebecca Batiste, Tianhao Zhao, Anand Shankar, Arvind Rao, Chao Chen, Dimitris Samaras, Tahsin Kurc, et~al.
\newblock Deep learning-based mapping of tumor infiltrating lymphocytes in whole slide images of 23 types of cancer.
\newblock \emph{Frontiers in Oncology}, 11:\penalty0 806603, 2022.

\bibitem[Aubreville et~al.(2023)Aubreville, Wilm, Stathonikos, Breininger, Donovan, Jabari, Veta, Ganz, Ammeling, van Diest, et~al.]{aubreville2023comprehensive}
Marc Aubreville, Frauke Wilm, Nikolas Stathonikos, Katharina Breininger, Taryn~A Donovan, Samir Jabari, Mitko Veta, Jonathan Ganz, Jonas Ammeling, Paul~J van Diest, et~al.
\newblock A comprehensive multi-domain dataset for mitotic figure detection.
\newblock \emph{Scientific Data}, 10\penalty0 (1):\penalty0 484, 2023.

\bibitem[Azizi et~al.(2023)Azizi, Culp, Freyberg, Mustafa, Baur, Kornblith, Chen, Tomasev, Mitrovi{\'c}, Strachan, et~al.]{azizi2023robust}
Shekoofeh Azizi, Laura Culp, Jan Freyberg, Basil Mustafa, Sebastien Baur, Simon Kornblith, Ting Chen, Nenad Tomasev, Jovana Mitrovi{\'c}, Patricia Strachan, et~al.
\newblock Robust and data-efficient generalization of self-supervised machine learning for diagnostic imaging.
\newblock \emph{Nature Biomedical Engineering}, 7\penalty0 (6):\penalty0 756--779, 2023.

\bibitem[Balestriero et~al.(2022)Balestriero, Bottou, and LeCun]{balestriero2022effects}
Randall Balestriero, Leon Bottou, and Yann LeCun.
\newblock The effects of regularization and data augmentation are class dependent.
\newblock In \emph{Advances in Neural Information Processing Systems}, 2022.

\bibitem[Balestriero et~al.(2023)Balestriero, Ibrahim, Sobal, Morcos, Shekhar, Goldstein, Bordes, Bardes, Mialon, Tian, Schwarzschild, Wilson, Geiping, Garrido, Fernandez, Bar, Pirsiavash, LeCun, and Goldblum]{Balestriero2023ACO}
Randall Balestriero, Mark Ibrahim, Vlad Sobal, Ari~S. Morcos, Shashank Shekhar, Tom Goldstein, Florian Bordes, Adrien Bardes, Gr{\'e}goire Mialon, Yuandong Tian, Avi Schwarzschild, Andrew~Gordon Wilson, Jonas Geiping, Quentin Garrido, Pierre Fernandez, Amir Bar, Hamed Pirsiavash, Yann LeCun, and Micah Goldblum.
\newblock A cookbook of self-supervised learning.
\newblock \emph{ArXiv}, abs/2304.12210, 2023.
\newblock URL \url{https://api.semanticscholar.org/CorpusID:258298825}.

\bibitem[Bardes et~al.(2021)Bardes, Ponce, and LeCun]{bardes2022vicreg}
Adrien Bardes, Jean Ponce, and Yann LeCun.
\newblock Vicreg: Variance-invariance-covariance regularization for self-supervised learning.
\newblock \emph{arXiv preprint arXiv:2105.04906}, 2021.

\bibitem[Bejnordi et~al.(2017)Bejnordi, Veta, Van~Diest, Van~Ginneken, Karssemeijer, Litjens, Van Der~Laak, Hermsen, Manson, Balkenhol, et~al.]{bejnordi2017diagnostic}
Babak~Ehteshami Bejnordi, Mitko Veta, Paul~Johannes Van~Diest, Bram Van~Ginneken, Nico Karssemeijer, Geert Litjens, Jeroen~AWM Van Der~Laak, Meyke Hermsen, Quirine~F Manson, Maschenka Balkenhol, et~al.
\newblock Diagnostic assessment of deep learning algorithms for detection of lymph node metastases in women with breast cancer.
\newblock \emph{JAMA}, 318\penalty0 (22):\penalty0 2199--2210, 2017.

\bibitem[Bordes et~al.(2023)Bordes, Balestriero, and Vincent]{bordes2023democratizing}
Florian Bordes, Randall Balestriero, and Pascal Vincent.
\newblock Towards democratizing joint-embedding self-supervised learning.
\newblock \emph{arXiv preprint arXiv:2303.01986}, 2023.

\bibitem[Borodachov et~al.(2019)Borodachov, Hardin, and Saff]{Borodachov2019DiscreteEO}
Sergiy~V Borodachov, Douglas~P Hardin, and Edward~B Saff.
\newblock \emph{Discrete energy on rectifiable sets}, volume~4.
\newblock Springer, 2019.

\bibitem[Campanella et~al.(2023)Campanella, Kwan, Fluder, Zeng, Stock, Veremis, Polydorides, Hedvat, Schoenfeld, Vanderbilt, et~al.]{campanella2023computational}
Gabriele Campanella, Ricky Kwan, Eugene Fluder, Jennifer Zeng, Aryeh Stock, Brandon Veremis, Alexandros~D Polydorides, Cyrus Hedvat, Adam Schoenfeld, Chad Vanderbilt, et~al.
\newblock Computational pathology at health system scale--self-supervised foundation models from three billion images.
\newblock \emph{arXiv preprint arXiv:2310.07033}, 2023.

\bibitem[Caron et~al.(2020)Caron, Misra, Mairal, Goyal, Bojanowski, and Joulin]{caron2020unsupervised}
Mathilde Caron, Ishan Misra, Julien Mairal, Priya Goyal, Piotr Bojanowski, and Armand Joulin.
\newblock Unsupervised learning of visual features by contrasting cluster assignments.
\newblock In \emph{Advances in Neural Information Processing Systems}, 2020.

\bibitem[Caron et~al.(2021)Caron, Touvron, Misra, J{\'e}gou, Mairal, Bojanowski, and Joulin]{caron2021emerging}
Mathilde Caron, Hugo Touvron, Ishan Misra, Herv{\'e} J{\'e}gou, Julien Mairal, Piotr Bojanowski, and Armand Joulin.
\newblock Emerging properties in self-supervised vision transformers.
\newblock In \emph{Proceedings of the IEEE/CVF International Conference on Computer Vision}, 2021.

\bibitem[Chen et~al.(2024)Chen, Ding, Lu, Williamson, Jaume, Song, Chen, Zhang, Shao, Shaban, et~al.]{chen2024towards}
Richard~J Chen, Tong Ding, Ming~Y Lu, Drew~FK Williamson, Guillaume Jaume, Andrew~H Song, Bowen Chen, Andrew Zhang, Daniel Shao, Muhammad Shaban, et~al.
\newblock Towards a general-purpose foundation model for computational pathology.
\newblock \emph{Nature Medicine}, pages 1--13, 2024.

\bibitem[Chen et~al.(2020{\natexlab{a}})Chen, Kornblith, Norouzi, and Hinton]{chen2020simple}
Ting Chen, Simon Kornblith, Mohammad Norouzi, and Geoffrey Hinton.
\newblock A simple framework for contrastive learning of visual representations.
\newblock In \emph{International Conference on Machine Learning}, 2020{\natexlab{a}}.

\bibitem[Chen and He(2021)]{chen2020exploring}
Xinlei Chen and Kaiming He.
\newblock Exploring simple siamese representation learning.
\newblock In \emph{Proceedings of the IEEE/CVF Conference on Computer Vision and Pattern Recognition}, 2021.

\bibitem[Chen et~al.(2020{\natexlab{b}})Chen, Fan, Girshick, and He]{chen2020improved}
Xinlei Chen, Haoqi Fan, Ross Girshick, and Kaiming He.
\newblock Improved baselines with momentum contrastive learning.
\newblock \emph{arXiv preprint arXiv:2003.04297}, 2020{\natexlab{b}}.

\bibitem[Chen et~al.(2021)Chen, Xie, and He]{chen2021empiricalstudytrainingselfsupervised}
Xinlei Chen, Saining Xie, and Kaiming He.
\newblock An empirical study of training self-supervised vision transformers.
\newblock In \emph{Proceedings of the IEEE/CVF International Conference on Computer Vision}, 2021.

\bibitem[Ciga et~al.(2022)Ciga, Xu, and Martel]{ciga2022self}
Ozan Ciga, Tony Xu, and Anne~Louise Martel.
\newblock Self supervised contrastive learning for digital histopathology.
\newblock \emph{Machine Learning with Applications}, 7:\penalty0 100198, 2022.

\bibitem[Darcet et~al.(2023)Darcet, Oquab, Mairal, and Bojanowski]{darcet2024visiontransformersneedregisters}
Timoth{\'e}e Darcet, Maxime Oquab, Julien Mairal, and Piotr Bojanowski.
\newblock Vision transformers need registers.
\newblock \emph{arXiv preprint arXiv:2309.16588}, 2023.

\bibitem[Dehghani et~al.(2023)Dehghani, Djolonga, Mustafa, Padlewski, Heek, Gilmer, Steiner, Caron, Geirhos, Alabdulmohsin, et~al.]{dehghani2023scalingvisiontransformers22}
Mostafa Dehghani, Josip Djolonga, Basil Mustafa, Piotr Padlewski, Jonathan Heek, Justin Gilmer, Andreas~Peter Steiner, Mathilde Caron, Robert Geirhos, Ibrahim Alabdulmohsin, et~al.
\newblock Scaling vision transformers to 22 billion parameters.
\newblock In \emph{International Conference on Machine Learning}, 2023.

\bibitem[Deng et~al.(2009)Deng, Dong, Socher, Li, Li, and Fei-Fei]{deng2009imagenet}
Jia Deng, Wei Dong, Richard Socher, Li-Jia Li, Kai Li, and Li~Fei-Fei.
\newblock Imagenet: A large-scale hierarchical image database.
\newblock In \emph{2009 IEEE Conference on Computer Vision and Pattern Recognition}, 2009.

\bibitem[Dippel et~al.(2024)Dippel, Feulner, Winterhoff, Schallenberg, Dernbach, Kunft, Tietz, Jurmeister, Horst, Ruff, et~al.]{dippel2024rudolfv}
Jonas Dippel, Barbara Feulner, Tobias Winterhoff, Simon Schallenberg, Gabriel Dernbach, Andreas Kunft, Stephan Tietz, Philipp Jurmeister, David Horst, Lukas Ruff, et~al.
\newblock Rudolfv: A foundation model by pathologists for pathologists.
\newblock \emph{arXiv preprint arXiv:2401.04079}, 2024.

\bibitem[Dosovitskiy et~al.(2020)Dosovitskiy, Beyer, Kolesnikov, Weissenborn, Zhai, Unterthiner, Dehghani, Minderer, Heigold, Gelly, et~al.]{dosovitskiy2020image}
Alexey Dosovitskiy, Lucas Beyer, Alexander Kolesnikov, Dirk Weissenborn, Xiaohua Zhai, Thomas Unterthiner, Mostafa Dehghani, Matthias Minderer, Georg Heigold, Sylvain Gelly, et~al.
\newblock An image is worth 16x16 words: Transformers for image recognition at scale.
\newblock \emph{arXiv preprint arXiv:2010.11929}, 2020.

\bibitem[Faryna et~al.(2021)Faryna, van~der Laak, and Litjens]{pmlr-v143-faryna21a}
Khrystyna Faryna, Jeroen van~der Laak, and Geert Litjens.
\newblock Tailoring automated data augmentation to h\&e-stained histopathology.
\newblock In \emph{Proceedings of the Fourth Conference on Medical Imaging with Deep Learning}, 2021.

\bibitem[Filiot et~al.(2023)Filiot, Ghermi, Olivier, Jacob, Fidon, Mac~Kain, Saillard, and Schiratti]{filiot2023scaling}
Alexandre Filiot, Ridouane Ghermi, Antoine Olivier, Paul Jacob, Lucas Fidon, Alice Mac~Kain, Charlie Saillard, and Jean-Baptiste Schiratti.
\newblock Scaling self-supervised learning for histopathology with masked image modeling.
\newblock \emph{medRxiv preprint 23292757}, 2023.

\bibitem[Grill et~al.(2020)Grill, Strub, Altch{\'e}, Tallec, Richemond, Buchatskaya, Doersch, Avila~Pires, Guo, Gheshlaghi~Azar, Piot, kavukcuoglu, Munos, and Valko]{grill2020bootstrap}
Jean-Bastien Grill, Florian Strub, Florent Altch{\'e}, Corentin Tallec, Pierre Richemond, Elena Buchatskaya, Carl Doersch, Bernardo Avila~Pires, Zhaohan Guo, Mohammad Gheshlaghi~Azar, Bilal Piot, koray kavukcuoglu, Remi Munos, and Michal Valko.
\newblock Bootstrap your own latent-a new approach to self-supervised learning.
\newblock In \emph{Advances in Neural Information Processing Systems}, 2020.

\bibitem[Gullapally et~al.(2023)Gullapally, Zhang, Mittal, Kartik, Srinivasan, Rose, Shenker, Juyal, Padigela, Biju, et~al.]{gullapally2023synthetic}
Sai~Chowdary Gullapally, Yibo Zhang, Nitin~Kumar Mittal, Deeksha Kartik, Sandhya Srinivasan, Kevin Rose, Daniel Shenker, Dinkar Juyal, Harshith Padigela, Raymond Biju, et~al.
\newblock Synthetic domain-targeted augmentation ({S-DOTA}) improves model generalization in digital pathology.
\newblock \emph{arXiv preprint arXiv:2305.02401}, 2023.

\bibitem[Han et~al.(2022)Han, Ye, and Zhan]{han2022augmentation}
Lu~Han, Han-Jia Ye, and De-Chuan Zhan.
\newblock Augmentation component analysis: Modeling similarity via the augmentation overlaps.
\newblock \emph{arXiv preprint arXiv:2206.00471}, 2022.

\bibitem[He et~al.(2022)He, Chen, Xie, Li, Doll{\'a}r, and Girshick]{he2021masked}
Kaiming He, Xinlei Chen, Saining Xie, Yanghao Li, Piotr Doll{\'a}r, and Ross Girshick.
\newblock Masked autoencoders are scalable vision learners.
\newblock In \emph{Proceedings of the IEEE/CVF Conference on Computer Vision and Pattern Recognition}, 2022.

\bibitem[Huang et~al.(2021)Huang, Yi, Zhao, and Jiang]{huang2021towards}
Weiran Huang, Mingyang Yi, Xuyang Zhao, and Zihao Jiang.
\newblock Towards the generalization of contrastive self-supervised learning.
\newblock \emph{arXiv preprint arXiv:2111.00743}, 2021.

\bibitem[Ilse et~al.(2018)Ilse, Tomczak, and Welling]{ilse2018attention}
Maximilian Ilse, Jakub Tomczak, and Max Welling.
\newblock Attention-based deep multiple instance learning.
\newblock In \emph{International conference on machine learning}, pages 2127--2136. PMLR, 2018.

\bibitem[Jaume et~al.(2024)Jaume, Doucet, Song, Lu, Almagro-P{\'e}rez, Wagner, Vaidya, Chen, Williamson, Kim, et~al.]{jaume2024hest}
Guillaume Jaume, Paul Doucet, Andrew~H Song, Ming~Y Lu, Cristina Almagro-P{\'e}rez, Sophia~J Wagner, Anurag~J Vaidya, Richard~J Chen, Drew~FK Williamson, Ahrong Kim, et~al.
\newblock Hest-1k: A dataset for spatial transcriptomics and histology image analysis.
\newblock \emph{arXiv preprint arXiv:2406.16192}, 2024.

\bibitem[Juyal et~al.(2024)Juyal, Padigela, Shah, Shenker, Harguindeguy, Liu, Martin, Zhang, Nercessian, Markey, et~al.]{juyal2024pluto}
Dinkar Juyal, Harshith Padigela, Chintan Shah, Daniel Shenker, Natalia Harguindeguy, Yi~Liu, Blake Martin, Yibo Zhang, Michael Nercessian, Miles Markey, et~al.
\newblock Pluto: Pathology-universal transformer.
\newblock \emph{arXiv preprint arXiv:2405.07905}, 2024.

\bibitem[Kaczmarzyk et~al.(2022)Kaczmarzyk, Abousamra, Kurc, Gupta, and Saltz]{kaczmarzyk10dataset}
JR~Kaczmarzyk, S~Abousamra, T~Kurc, R~Gupta, and J~Saltz.
\newblock Dataset for tumor infiltrating lymphocyte classification (304,097 image patches from tcga), 2022.
\newblock \emph{URL https://doi. org/10}, 5281, 2022.

\bibitem[Kang et~al.(2023)Kang, Song, Park, Yoo, and Pereira]{kang2023benchmarking}
Mingu Kang, Heon Song, Seonwook Park, Donggeun Yoo, and S{\'e}rgio Pereira.
\newblock Benchmarking self-supervised learning on diverse pathology datasets.
\newblock In \emph{Proceedings of the IEEE/CVF Conference on Computer Vision and Pattern Recognition}, 2023.

\bibitem[Kaplan et~al.(2020)Kaplan, McCandlish, Henighan, Brown, Chess, Child, Gray, Radford, Wu, and Amodei]{kaplan2020scaling}
Jared Kaplan, Sam McCandlish, Tom Henighan, Tom~B Brown, Benjamin Chess, Rewon Child, Scott Gray, Alec Radford, Jeffrey Wu, and Dario Amodei.
\newblock Scaling laws for neural language models.
\newblock \emph{arXiv preprint arXiv:2001.08361}, 2020.

\bibitem[Kather et~al.(2018)Kather, Halama, and Marx]{kather_jakob_nikolas_2018_1214456}
Jakob~Nikolas Kather, Niels Halama, and Alexander Marx.
\newblock 100,000 histological images of human colorectal cancer and healthy tissue.
\newblock \emph{Zenodo}, April 2018.
\newblock \doi{10.5281/zenodo.1214456}.
\newblock URL \url{https://doi.org/10.5281/zenodo.1214456}.

\bibitem[Koh et~al.(2021)Koh, Sagawa, Marklund, Xie, Zhang, Balsubramani, Hu, Yasunaga, Phillips, Gao, et~al.]{koh2021wilds}
Pang~Wei Koh, Shiori Sagawa, Henrik Marklund, Sang~Michael Xie, Marvin Zhang, Akshay Balsubramani, Weihua Hu, Michihiro Yasunaga, Richard~Lanas Phillips, Irena Gao, et~al.
\newblock Wilds: A benchmark of in-the-wild distribution shifts.
\newblock In \emph{International Conference on Machine Learning}, 2021.

\bibitem[Kumar et~al.(2023)Kumar, Dehghani, and Houlsby]{kumar2023dualpatchnorm}
Manoj Kumar, Mostafa Dehghani, and Neil Houlsby.
\newblock Dual patchnorm.
\newblock \emph{arXiv preprint arXiv:2302.01327}, 2023.

\bibitem[Lab(2024)]{hest_github}
Mahmood Lab.
\newblock Hest.
\newblock \url{https://github.com/mahmoodlab/HEST/tree/v1.0.2}, 2024.

\bibitem[Molybog et~al.(2023)Molybog, Albert, Chen, DeVito, Esiobu, Goyal, Koura, Narang, Poulton, Silva, et~al.]{molybog2023theoryadaminstabilitylargescale}
Igor Molybog, Peter Albert, Moya Chen, Zachary DeVito, David Esiobu, Naman Goyal, Punit~Singh Koura, Sharan Narang, Andrew Poulton, Ruan Silva, et~al.
\newblock A theory on adam instability in large-scale machine learning.
\newblock \emph{arXiv preprint arXiv:2304.09871}, 2023.

\bibitem[Moutakanni et~al.(2024)Moutakanni, Oquab, Szafraniec, Vakalopoulou, and Bojanowski]{moutakanni2024dontneeddataaugmentationselfsupervised}
Th{\'e}o Moutakanni, Maxime Oquab, Marc Szafraniec, Maria Vakalopoulou, and Piotr Bojanowski.
\newblock You don't need data-augmentation in self-supervised learning.
\newblock \emph{arXiv preprint arXiv:2406.09294}, 2024.

\bibitem[Nechaev et~al.(2024)Nechaev, Pchelnikov, and Ivanova]{nechaev2024hibou}
Dmitry Nechaev, Alexey Pchelnikov, and Ekaterina Ivanova.
\newblock Hibou: A family of foundational vision transformers for pathology.
\newblock \emph{arXiv preprint arXiv:2406.05074}, 2024.

\bibitem[OpenAI(2023)]{openai_gpt-4_2023}
OpenAI.
\newblock {GPT}-4 {Technical} {Report}.
\newblock \emph{\textup{Preprint at https://doi.org/10.48550/arXiv.2303.08774}}, 2023.

\bibitem[Oquab et~al.(2023)Oquab, Darcet, Moutakanni, Vo, Szafraniec, Khalidov, Fernandez, Haziza, Massa, El-Nouby, et~al.]{oquab2024dinov2}
Maxime Oquab, Timoth{\'e}e Darcet, Th{\'e}o Moutakanni, Huy Vo, Marc Szafraniec, Vasil Khalidov, Pierre Fernandez, Daniel Haziza, Francisco Massa, Alaaeldin El-Nouby, et~al.
\newblock Dinov2: Learning robust visual features without supervision.
\newblock \emph{arXiv preprint arXiv:2304.07193}, 2023.

\bibitem[Sablayrolles et~al.(2018)Sablayrolles, Douze, Schmid, and J{\'e}gou]{sablayrolles2019spreading}
Alexandre Sablayrolles, Matthijs Douze, Cordelia Schmid, and Herv{\'e} J{\'e}gou.
\newblock Spreading vectors for similarity search.
\newblock \emph{arXiv preprint arXiv:1806.03198}, 2018.

\bibitem[Saillard et~al.(2024)Saillard, Jenatton, Llinares-López, Mariet, Cahané, Durand, and Vert]{hoptimus0}
Charlie Saillard, Rodolphe Jenatton, Felipe Llinares-López, Zelda Mariet, David Cahané, Eric Durand, and Jean-Philippe Vert.
\newblock H-optimus-0, 2024.
\newblock URL \url{https://github.com/bioptimus/releases/tree/main/models/h-optimus/v0}.

\bibitem[Saltz et~al.(2018)Saltz, Gupta, Hou, Kurc, Singh, Nguyen, Samaras, Shroyer, Zhao, Batiste, et~al.]{saltz2018spatial}
Joel Saltz, Rajarsi Gupta, Le~Hou, Tahsin Kurc, Pankaj Singh, Vu~Nguyen, Dimitris Samaras, Kenneth~R Shroyer, Tianhao Zhao, Rebecca Batiste, et~al.
\newblock Spatial organization and molecular correlation of tumor-infiltrating lymphocytes using deep learning on pathology images.
\newblock \emph{Cell Reports}, 23\penalty0 (1):\penalty0 181--193, 2018.

\bibitem[Shen et~al.(2022)Shen, Luo, Shen, and Ke]{shen2022randstainna}
Yiqing Shen, Yulin Luo, Dinggang Shen, and Jing Ke.
\newblock Randstainna: Learning stain-agnostic features from histology slides by bridging stain augmentation and normalization.
\newblock In \emph{International Conference on Medical Image Computing and Computer-Assisted Intervention}, 2022.

\bibitem[Sun et~al.(2017)Sun, Shrivastava, Singh, and Gupta]{sun2017revisiting}
Chen Sun, Abhinav Shrivastava, Saurabh Singh, and Abhinav Gupta.
\newblock Revisiting unreasonable effectiveness of data in deep learning era.
\newblock In \emph{Proceedings of the IEEE International Conference on Computer Vision}, 2017.

\bibitem[Tellez et~al.(2019)Tellez, Litjens, B{\'a}ndi, Bulten, Bokhorst, Ciompi, and Van Der~Laak]{tellez2019quantifying}
David Tellez, Geert Litjens, P{\'e}ter B{\'a}ndi, Wouter Bulten, John-Melle Bokhorst, Francesco Ciompi, and Jeroen Van Der~Laak.
\newblock Quantifying the effects of data augmentation and stain color normalization in convolutional neural networks for computational pathology.
\newblock \emph{Medical image analysis}, 58:\penalty0 101544, 2019.

\bibitem[Veeling et~al.(2018)Veeling, Linmans, Winkens, Cohen, and Welling]{veeling2018rotation}
Bastiaan~S Veeling, Jasper Linmans, Jim Winkens, Taco Cohen, and Max Welling.
\newblock Rotation equivariant cnns for digital pathology.
\newblock In \emph{Medical Image Computing and Computer Assisted Intervention}, 2018.

\bibitem[Vorontsov et~al.(2024)Vorontsov, Bozkurt, Casson, Shaikovski, Zelechowski, Severson, Zimmermann, Hall, Tenenholtz, Fusi, et~al.]{vorontsov2024virchow}
Eugene Vorontsov, Alican Bozkurt, Adam Casson, George Shaikovski, Michal Zelechowski, Kristen Severson, Eric Zimmermann, James Hall, Neil Tenenholtz, Nicolo Fusi, et~al.
\newblock A foundation model for clinical-grade computational pathology and rare cancers detection.
\newblock \emph{Nature Medicine}, pages 1--12, 2024.

\bibitem[Vrabac et~al.(2021)Vrabac, Smit, Rojansky, Natkunam, Advani, Ng, Fernandez-Pol, and Rajpurkar]{vrabac2021dlbcl}
Damir Vrabac, Akshay Smit, Rebecca Rojansky, Yasodha Natkunam, Ranjana~H Advani, Andrew~Y Ng, Sebastian Fernandez-Pol, and Pranav Rajpurkar.
\newblock Dlbcl-morph: morphological features computed using deep learning for an annotated digital dlbcl image set.
\newblock \emph{Scientific Data}, 8\penalty0 (1):\penalty0 135, 2021.

\bibitem[Wang and Isola(2020)]{wang2022understanding}
Tongzhou Wang and Phillip Isola.
\newblock Understanding contrastive representation learning through alignment and uniformity on the hypersphere.
\newblock In \emph{International Conference on Machine Learning}, 2020.

\bibitem[Wang et~al.(2022)Wang, Yang, Zhang, Wang, Zhang, Yang, Huang, and Han]{wang2022transformer}
Xiyue Wang, Sen Yang, Jun Zhang, Minghui Wang, Jing Zhang, Wei Yang, Junzhou Huang, and Xiao Han.
\newblock Transformer-based unsupervised contrastive learning for histopathological image classification.
\newblock \emph{Medical image analysis}, 81:\penalty0 102559, 2022.

\bibitem[Wei et~al.(2021)Wei, Suriawinata, Ren, Liu, Lisovsky, Vaickus, Brown, Baker, Tomita, Torresani, et~al.]{wei2021petri}
Jerry Wei, Arief Suriawinata, Bing Ren, Xiaoying Liu, Mikhail Lisovsky, Louis Vaickus, Charles Brown, Michael Baker, Naofumi Tomita, Lorenzo Torresani, et~al.
\newblock A petri dish for histopathology image analysis.
\newblock In \emph{Artificial Intelligence in Medicine: 19th International Conference on Artificial Intelligence in Medicine}, 2021.

\bibitem[Weinstein et~al.(2013)Weinstein, Collisson, Mills, Shaw, Ozenberger, Ellrott, Shmulevich, Sander, and Stuart]{weinstein2013cancer}
John~N Weinstein, Eric~A Collisson, Gordon~B Mills, Kenna~R Shaw, Brad~A Ozenberger, Kyle Ellrott, Ilya Shmulevich, Chris Sander, and Joshua~M Stuart.
\newblock The cancer genome atlas pan-cancer analysis project.
\newblock \emph{Nature Genetics}, 45\penalty0 (10):\penalty0 1113--1120, 2013.

\bibitem[Wortsman et~al.(2023)Wortsman, Dettmers, Zettlemoyer, Morcos, Farhadi, and Schmidt]{wortsman2023stablelowprecisiontraininglargescale}
Mitchell Wortsman, Tim Dettmers, Luke Zettlemoyer, Ari Morcos, Ali Farhadi, and Ludwig Schmidt.
\newblock Stable and low-precision training for large-scale vision-language models.
\newblock In \emph{Advances in Neural Information Processing Systems}, 2023.

\bibitem[Xu et~al.(2024)Xu, Usuyama, Bagga, Zhang, Rao, Naumann, Wong, Gero, González, Gu, Xu, Wei, Wang, Ma, Wei, Yang, Li, Gao, Rosemon, Bower, Lee, Weerasinghe, Wright, Robicsek, Piening, Bifulco, Wang, and Poon]{Xu2024}
Hanwen Xu, Naoto Usuyama, Jaspreet Bagga, Sheng Zhang, Rajesh Rao, Tristan Naumann, Cliff Wong, Zelalem Gero, Javier González, Yu~Gu, Yanbo Xu, Mu~Wei, Wenhui Wang, Shuming Ma, Furu Wei, Jianwei Yang, Chunyuan Li, Jianfeng Gao, Jaylen Rosemon, Tucker Bower, Soohee Lee, Roshanthi Weerasinghe, Bill~J. Wright, Ari Robicsek, Brian Piening, Carlo Bifulco, Sheng Wang, and Hoifung Poon.
\newblock A whole-slide foundation model for digital pathology from real-world data.
\newblock \emph{Nature}, 630\penalty0 (8015):\penalty0 181–188, 2024.

\bibitem[Yeh et~al.(2022)Yeh, Hong, Hsu, Liu, Chen, and LeCun]{yeh2022decoupled}
Chun-Hsiao Yeh, Cheng-Yao Hong, Yen-Chi Hsu, Tyng-Luh Liu, Yubei Chen, and Yann LeCun.
\newblock Decoupled contrastive learning.
\newblock In \emph{European Conference on Computer Vision}, 2022.

\bibitem[Zbontar et~al.(2021)Zbontar, Jing, Misra, LeCun, and Deny]{zbontar2021barlow}
Jure Zbontar, Li~Jing, Ishan Misra, Yann LeCun, and St{\'e}phane Deny.
\newblock Barlow twins: Self-supervised learning via redundancy reduction.
\newblock In \emph{International Conference on Machine Learning}, 2021.

\bibitem[Zhai et~al.(2022)Zhai, Kolesnikov, Houlsby, and Beyer]{zhai2022scaling}
Xiaohua Zhai, Alexander Kolesnikov, Neil Houlsby, and Lucas Beyer.
\newblock Scaling vision transformers.
\newblock In \emph{Proceedings of the IEEE/CVF Conference on Computer Vision and Pattern Recognition}, 2022.

\bibitem[Zhou et~al.(2021)Zhou, Wei, Wang, Shen, Xie, Yuille, and Kong]{zhou2021ibot}
Jinghao Zhou, Chen Wei, Huiyu Wang, Wei Shen, Cihang Xie, Alan Yuille, and Tao Kong.
\newblock ibot: Image bert pre-training with online tokenizer.
\newblock \emph{arXiv preprint arXiv:2111.07832}, 2021.

\end{thebibliography}

\newpage
\appendix
\setcounter{table}{0}
\setcounter{figure}{0}
\renewcommand{\thetable}{A\arabic{table}}
\renewcommand{\thefigure}{A\arabic{figure}}
\section{Appendix}
\subsection{Foundation models in CPath}

\begin{table}[h]
\centering
\begin{tabular}{rcrrcrc}
\toprule
\multirow{2}[0]{*}{Model} & \multirow{2}[0]{*}{Data source} & \multicolumn{2}{c}{Data size}      & \multirow{2}[0]{*}{Model architecture} & \multirow{2}[0]{*}{Model size} & \multirow{2}[0]{*}{Objective function} \\
\cmidrule{3-4}
                       &                              & WSI                       & Tiles &                                     &                             &                                     \\ \toprule
   Virchow2G & MSKCC & 3.1M & 1.9B & ViT-G & 1.9B & DINOv2 \\ \midrule
   Virchow2G Mini & MSKCC & 3.1M & 0.9B & ViT-S & 22M & DINOv2 \\ \midrule
  Virchow2 & MSKCC & 3.1M & 1.7B & ViT-H & 632M & DINOv2 \\ \midrule 
Virchow    ~\cite{vorontsov2024virchow}      & MSKCC                        & 1.5M &   2B      & ViT-H                               & 632M                        & DINOv2                              \\ \midrule
Hibou~\cite{nechaev2024hibou} & Proprietary & 1.1M & 1.2B & ViT-L & 307M & DINOv2  \\   \midrule 
  H-optimus-0~\cite{hoptimus0} & Proprietary & 500K & & ViT-g & 1.1B & DINOv2 \\   \midrule
  Campanella et al.~\cite{campanella2023computational}       & Mount Sinai                  & 400K & 3B      & ViT-S                               & 22M                         & DINO, MAE                                \\ \midrule
      Prov-GigaPath~\cite{Xu2024} & Providence & 170K & 1.3B & ViT-g & 1.1B & DINOv2 \\   \midrule     
   PLUTO~\cite{juyal2024pluto} & Proprietary & 158K & 195M & ViT-S & 22M &   {\begin{tabular}[c]{@{}c@{}}DINOv2 + MAE \\ + Fourier\end{tabular}}\\ \midrule 
RudolfV~\cite{dippel2024rudolfv} & TCGA + Properitary & 103K & 750M & ViT-L & 307M & DINOv2 \\ \midrule
UNI~\cite{chen2024towards}                    & Mass-100K                    & 100K & 100M    & ViT-L                               & 307M                        & DINOv2                              \\ \midrule
Lunit~\cite{kang2023benchmarking}                  & TCGA + TULIP                 & 37K  & 33M     & ViT-S                               & 22M                         & Various                              \\ \midrule
CTransPath~\cite{wang2022transformer}             & TCGA + PAIP                  & 32K  & 15M     & Swin Transformer                     & 28M                         & MoCoV3                                                    \\ \midrule
Remedis~\cite{azizi2023robust}                & TCGA                         & 29K  & 50M     & ResNet-152                          & 232M                        & SimCLR                              \\ \midrule
Ciga et al.~\cite{ciga2022self}            & TCGA + CPTAC ++              & 25K  & 4.2M    & ResNet                              & 11-45M                      & SimCLR                              \\ \midrule
Phikon~\cite{filiot2023scaling}                 & TCGA                         & 6K   & 43M     & ViT-B                               & 86M                         & iBOT                                \\ \midrule
 \bottomrule
\end{tabular}
\caption{Summary of proposed foundation models in computational pathology highlighting the size of the training data, size of the model architecture, and training objective. Works are ordered based on the number of training WSI.}
\label{tab:supp_hist_FM}
\end{table}

\clearpage

\subsection{Hyperparameters}

\begin{table*}[h]
    \centering
    \resizebox{6.4cm}{!}{
    \begin{tabular}{@{}ll@{}}
    \toprule
    Virchow2 Hyperparameter                & Value            \\
    \midrule
    Vision transformer                                         \\
    Patch size                              & 14               \\
    Embedding dimension                     & 1280             \\
    Layers                                  & 32               \\
    Heads                                   & 16               \\
    MLP ratio                               & 4                \\
    MLP activation                          & SwiGLU           \\
    MLP bias                                & True             \\
    QKV bias                                & True             \\
    Registers                               & 4                \\
    QK normalization                        & False            \\
    Dual PatchNorm                          & False            \\
    \midrule
    Projection heads                                           \\
    Shared heads                            & False            \\
    Layers                                  & 3                \\
    Bottleneck dimension                    & 384              \\
    Hidden dimension                        & 2048             \\
    \midrule
    Loss functions                                             \\
    Sinkhorn centering                      & True             \\  
    Student temperature                     & 0.1              \\
    Teacher temperature                     & (0.07, 0.04)     \\
    Prototypes                              & 131072           \\
    Regularizer                             & KDE              \\
    Regularizer parameter                   & 5                \\
    Regularizer weight                      & 0.05             \\
    \midrule
    Augmentations \\
    Tile context size                       & 392             \\
    Global view size                        & 224             \\
    Local view size                         & 98              \\
    Global crops                            & 2               \\
    Local crops                             & 8               \\
    Method                                  & ECT             \\
    Aspect ratio range                      & (0.95, 1.05)    \\
    Scale range                             & (0.9, 1.1)      \\
    Solarization                            & True            \\
    Vertical flips                          & True            \\
    \midrule
    Optimizer                               & AdamW           \\
    Optimizer momentum                      & (0.9, 0.999)    \\
    Optimizer epsilon                       & $1\times10^{-8}$\\
    Optimizer weight decay                  & (0.04, 0.2)     \\
    Learning rate                           & $2\times10^{-4}$\\
    Learning rate schedule                  & Cosine          \\
    Learning rate scaling                   & Square root-1024\\ 
    Teacher momentum                        & (0.994, 1.0)    \\
    Batch size                              & 4096            \\
    Gradient clipping norm                  & 3.0             \\
    Drop rate                               & 0.4             \\
    Precision                               & FP16            \\
    \bottomrule
    \end{tabular}%
    }
    \caption{Virchow2 adapted hyperparameters from DINOv2 trained on 512 NVIDIA V100 32GB GPUs. Default augmentations are not included and are left unchanged.}
    \label{tab:v2-hyperparameters}%
\end{table*}%

\begin{table*}[h]
    \centering
    \resizebox{6.4cm}{!}{
    \begin{tabular}{@{}ll@{}}
    \toprule
    Virchow2G hyperparameter               & Value            \\
    \midrule
    Vision transformer                                         \\
    Patch size                              & 14               \\
    Embedding dimension                     & 1792             \\
    Layers                                  & 48               \\
    Heads                                   & 28               \\
    MLP ratio                               & 4                \\
    MLP activation                          & SwiGLU           \\
    MLP bias                                & True             \\
    QKV bias                                & True             \\
    Registers                               & 8                \\
    QK normalization                        & True             \\
    Dual PatchNorm                          & True             \\
    \midrule
    Projection heads                                           \\
    Shared heads                            & False            \\
    Layers                                  & 3                \\
    Bottleneck dimension                    & 384              \\
    Hidden dimension                        & 2048             \\
    \midrule
    Loss functions                                             \\
    Sinkhorn centering                      & True             \\  
    Student temperature                     & 0.1              \\
    Teacher temperature                     & 0.04             \\
    Prototypes                              & 131072           \\
    Regularizer                             & KDE              \\
    Regularizer parameter                   & 5                \\
    Regularizer weight                      & 0.05             \\
    \midrule
    Augmentations \\
    Tile context size                       & 392             \\
    Global view size                        & 224             \\
    Local view size                         & 98              \\
    Global crops                            & 2               \\
    Local crops                             & 8               \\
    Method                                  & ECT             \\
    Aspect ratio range                      & (0.95, 1.05)    \\
    Scale range                             & (0.9, 1.1)      \\
    Solarization                            & False           \\
    Vertical flips                          & True            \\
    \midrule
    Optimizer                               & StableAdamW     \\
    Optimizer momentum                      & (0.9, 0.95)     \\
    Optimizer epsilon                       & $1\times10^{-6}$\\
    Optimizer weight decay                  & (0.04, 0.2)     \\
    Learning rate                           & $1\times10^{-4}$\\
    Learning rate schedule                  & Cosine          \\
    Learning rate scaling                   & Square Root-1024\\ 
    Teacher momentum                        & (0.994, 1.0)    \\
    Batch size                              & 3072            \\
    Gradient clipping norm                  & 3.0             \\
    Drop rate                               & 0.4             \\
    Precision                               & FP16            \\
    \bottomrule
    \end{tabular}%
    }
    \caption{Virchow2G adapted hyperparameters from DINOv2 trained on 512 NVIDIA V100 32GB GPUs. Default augmentations are not included and are left unchanged.}
    \label{tab:v2g-hyperparameters}%
\end{table*}%

\begin{table*}[h]
    \centering
    \resizebox{6.4cm}{!}{
    \begin{tabular}{@{}ll@{}}
    \toprule
    Virchow2G Mini hyperparameter           & Value            \\
    \midrule
    Vision transformer                                         \\
    Patch size                              & 14               \\
    Embedding dimension                     & 384              \\
    Layers                                  & 12               \\
    Heads                                   & 6                \\
    MLP ratio                               & 4                \\
    MLP activation                          & GELU             \\
    MLP bias                                & True             \\
    QKV bias                                & True             \\
    Registers                               & 4                \\
    QK normalization                        & False            \\
    Dual PatchNorm                          & False            \\
    \midrule
    Projection heads                                           \\
    Shared heads                            & False            \\
    Layers                                  & 3                \\
    Bottleneck dimension                    & 384              \\
    Hidden dimension                        & 2048             \\
    \midrule
    Loss functions                                             \\
    Sinkhorn centering                      & True             \\  
    Student temperature                     & 0.1              \\
    Teacher temperature                     & 0.04             \\
    Prototypes                              & 131072           \\
    Regularizer                             & KDE              \\
    Regularizer parameter                   & 5                \\
    Regularizer weight                      & 0.05             \\
    \midrule
    Augmentations \\
    Tile context size                       & 392             \\
    Global view size                        & 224             \\
    Local view size                         & 98              \\
    Global crops                            & 2               \\
    Local crops                             & 8               \\
    Method                                  & ECT             \\
    Aspect ratio range                      & (0.95, 1.05)    \\
    Scale range                             & (0.9, 1.1)      \\
    Solarization                            & False           \\
    Vertical flips                          & True            \\
    \midrule
    Optimizer                               & StableAdamW     \\
    Optimizer momentum                      & (0.9, 0.98)     \\
    Optimizer epsilon                       & $1\times10^{-6}$\\
    Optimizer weight decay                  & (0.04, 0.2)     \\
    Learning rate                           & $2\times10^{-4}$\\
    Learning rate schedule                  & Cosine          \\
    Learning rate scaling                   & Square Root-1024\\ 
    Student EMA copy momentum               & 0.994           \\
    Batch size                              & 2048            \\
    Gradient clipping norm                  & 3.0             \\
    Drop rate                               & 0.0             \\
    Precision                               & FP16            \\
    \bottomrule
    \end{tabular}%
    }
    \caption{Virchow2G Mini adapted hyperparameters from DINOv2 trained on 256 NVIDIA V100 32GB GPUs. Default augmentations are not included and are left unchanged.}
    \label{tab:v2gmini-hyperparameters}%
\end{table*}%

\clearpage
\subsection{Detailed ablation results}
\label{sec:appendix_ablation}

Here we present the detailed results of the ablation study. Tables~\ref{app-tab:id} and~\ref{app-tab:ood}, presented the weighted F1 score for the in-domain and out-of-domain tile tasks respectively. Note all metrics have been multiplied by 100 for improved readability. The bolded numbers indicate the best performing model; in cases where multiple results are bolded, the results are not statistically significantly different at $\alpha = 0.05$ as measured by McNemar's test. We do not apply statistical testing to the average performance results and therefore do not bold any values. Tables~\ref{tab:ablation_id_cls_only} and~\ref{tab:abalation_ood_cls_only} present the same analysis using only the [CLS] token as the embedding. The [CLS] token results have similar trends to those described in the main text. In general, the experiments with fewer modifications have a boost in performance when using the concatenated embedding but that delta decreases in the case where +ECT, +KDE or all three proposals are applied.

\begin{table}[h]
\centering
\small
\begin{tabular}{lllll}
\hline
 & \begin{tabular}[c]{@{}c@{}}PanMSK\\ 20x\end{tabular} & \begin{tabular}[c]{@{}c@{}}PanMSK\\ 10x\end{tabular} & \begin{tabular}[c]{@{}c@{}}PanMSK\\ 5x\end{tabular} & \begin{tabular}[c]{@{}c@{}}ID\\ Average\end{tabular} \\ \hline \hline
Standard DINOv2 & 86.0   & 88.8  & 89.0  & 87.9 \\
\quad+ECT
 & 86.5  {\color[HTML]{32CB00} $\uparrow$ 0.5}  & 89.4   {\color[HTML]{32CB00} $\uparrow$ 0.6}  & 89.9  {\color[HTML]{32CB00} $\uparrow$ 0.9} & 88.6 {\color[HTML]{32CB00} $\uparrow$ 0.7}  \\
 \quad+KDE
 & 88.0  {\color[HTML]{32CB00} $\uparrow$ 2.0}  & 90.1   {\color[HTML]{32CB00} $\uparrow$ 1.3}  & 90.0  {\color[HTML]{32CB00} $\uparrow$ 1.1} & 89.4 {\color[HTML]{32CB00} $\uparrow$ 1.5}  \\                   \quad-SOL
 & 87.1 {\color[HTML]{32CB00} $\uparrow$ 1.1}  & 89.2   {\color[HTML]{32CB00} $\uparrow$ 0.3}  & 89.0  --  & 89.4 {\color[HTML]{32CB00} $\uparrow$ 1.5}  \\   
 \quad+ECT,+KDE
 & 89.6 {\color[HTML]{32CB00} $\uparrow$ 3.6}  & 92.1   {\color[HTML]{32CB00} $\uparrow$ 3.2}  & 92.2 {\color[HTML]{32CB00} $\uparrow$ 3.2} & 91.3 {\color[HTML]{32CB00} $\uparrow$ 3.4}  \\   
  \quad+ECT,+KDE,-SOL
 & \textbf{89.9} {\color[HTML]{32CB00} $\uparrow$ 3.9}  & \textbf{92.4}   {\color[HTML]{32CB00} $\uparrow$ 3.6}  & \textbf{93.3} {\color[HTML]{32CB00} $\uparrow$ 4.3} & 91.9 {\color[HTML]{32CB00} $\uparrow$ 4.0}  \\   \hline
\end{tabular}
\caption{Weighted F1 score (multiplied by 100) for the in-domain tasks. ECT is the proposed extended context translation, KDE is the proposed regularization function and SOL refers to the removal of solarization.}
\label{app-tab:id}
\end{table}

\begin{table}[h]
\centering
\small
\begin{tabular}{lllllll} \hline
 & PCam & CRC  &  TILS & MHIST & MIDOG & \begin{tabular}[c]{@{}c@{}}OOD\\ Average\end{tabular} \\ \hline \hline
Standard DINOv2                       & 83.6 & 94.0 &  92.5 & \textbf{79.4}  & 63.0  & 82.5      \\
\quad+ECT                           & 84.9 {\color[HTML]{32CB00} $\uparrow$ 1.3} & 93.3 {\color[HTML]{CB0000} $\downarrow$ 0.7}  & 92.6 {\color[HTML]{32CB00} $\uparrow$ 0.1} & 76.2 {\color[HTML]{CB0000} $\downarrow$ 3.2}  & 63.6 {\color[HTML]{32CB00} $\uparrow$ 0.6}  & 82.1 {\color[HTML]{CB0000} $\downarrow$ 0.5}      \\
\quad+KDE                           & 84.1 {\color[HTML]{32CB00} $\uparrow$ 0.6} & 94.9 {\color[HTML]{32CB00} $\uparrow$ 0.9} &  \textbf{93.3} {\color[HTML]{32CB00} $\uparrow$ 0.8} & \textbf{78.3}  {\color[HTML]{CB0000} $\downarrow$ 1.0}  & \textbf{66.5} {\color[HTML]{32CB00} $\uparrow$ 3.5} & 83.4   {\color[HTML]{32CB00} $\uparrow$ 1.0}     \\
\quad-SOL                           & 84.8 {\color[HTML]{32CB00} $\uparrow$ 1.3} & 93.8 {\color[HTML]{CB0000} $\downarrow$ 0.2}  & 92.7 {\color[HTML]{32CB00} $\uparrow$ 0.2} & 77.2 {\color[HTML]{CB0000} $\downarrow$ 2.2}   & 65.9 {\color[HTML]{32CB00} $\uparrow$ 2.9} & 82.9  {\color[HTML]{32CB00} $\uparrow$ 0.4}      \\
\quad+ECT, +KDE                     & \textbf{86.7} {\color[HTML]{32CB00} $\uparrow$ 3.1} & 95.3 {\color[HTML]{32CB00} $\uparrow$ 1.3} & 93.0 {\color[HTML]{32CB00} $\uparrow$ 0.5} & \textbf{80.3} {\color[HTML]{32CB00} $\uparrow$ 0.9} & \textbf{66.5} {\color[HTML]{32CB00} $\uparrow$ 3.6} & 84.4 {\color[HTML]{32CB00} $\uparrow$ 1.9}      \\
\quad+ECT, +KDE, -SOL               & \textbf{86.7} {\color[HTML]{32CB00} $\uparrow$ 3.2} & \textbf{95.8}  {\color[HTML]{32CB00} $\uparrow$ 1.8}  & \textbf{93.2} {\color[HTML]{32CB00} $\uparrow$ 0.7} & \textbf{79.2} {\color[HTML]{CB0000} $\downarrow$ 0.2}  & \textbf{66.3} {\color[HTML]{32CB00} $\uparrow$ 3.4} & 84.3  {\color[HTML]{32CB00} $\uparrow$ 1.8}    \\ \hline
\end{tabular}
\caption{Weighted F1 score (multiplied by 100) for the out-of-domain tasks. ECT is the proposed extended context translation, KDE is the proposed regularization function and SOL refers to the removal of solarization.}
\label{app-tab:ood}
\end{table}

\begin{table}[h]
\centering
\small
\begin{tabular}{lllll}
\hline
 & \begin{tabular}[c]{@{}c@{}}PanMSK\\ 20x\end{tabular} & \begin{tabular}[c]{@{}c@{}}PanMSK\\ 10x\end{tabular} & \begin{tabular}[c]{@{}c@{}}PanMSK\\ 5x\end{tabular} & \begin{tabular}[c]{@{}c@{}}ID\\ Average\end{tabular} \\ \hline \hline
Standard DINOv2   & 83.2   & 86.0  & 86.3    & 85.1    \\
\quad+ECT         & 82.7 {\color[HTML]{CB0000} $\downarrow$ 0.4}   & 86.1 {\color[HTML]{32CB00} $\uparrow$ 0.1}  & 85.9 {\color[HTML]{CB0000} $\downarrow$ 0.3}   & 84.9 {\color[HTML]{CB0000} $\downarrow$ 0.2}   \\
\quad+KDE       & 87.4 {\color[HTML]{32CB00} $\uparrow$ 4.2} & 89.8  {\color[HTML]{32CB00} $\uparrow$ 3.8} & 89.6 {\color[HTML]{32CB00} $\uparrow$ 3.3}    & 88.9  {\color[HTML]{32CB00} $\uparrow$ 3.8}    \\
\quad-SOL       & 84.4 {\color[HTML]{32CB00} $\uparrow$ 1.2} & 87.0  {\color[HTML]{32CB00} $\uparrow$ 1.0} & 86.9 {\color[HTML]{32CB00} $\uparrow$ 0.6}    & 86.1  {\color[HTML]{32CB00} $\uparrow$ 1.0}  \\
\quad+ECT,+KDE & 89.3 {\color[HTML]{32CB00} $\uparrow$ 6.2}  & 91.9  {\color[HTML]{32CB00} $\uparrow$ 5.8}  & 92.1 {\color[HTML]{32CB00} $\uparrow$ 5.8}    & 91.1  {\color[HTML]{32CB00} $\uparrow$ 6.0}  \\
\quad+ECT, +KDE, -SOL  & \textbf{89.7} {\color[HTML]{32CB00} $\uparrow$ 6.6}  & \textbf{92.4} {\color[HTML]{32CB00} $\uparrow$ 6.4}  & \textbf{93.1} {\color[HTML]{32CB00} $\uparrow$ 6.9} & 91.8 {\color[HTML]{32CB00} $\uparrow$ 6.7} \\                                          \hline               
\end{tabular}
\caption{Weighted F1 score (multiplied by 100) for the in-domain tasks where the embedding is [CLS] only. ECT is the proposed extended context translation, KDE is the proposed regularization function and SOL refers to the removal of solarization.}
\label{tab:ablation_id_cls_only}
\end{table}

\begin{table}[h]
\small
\centering
\begin{tabular}{lllllllll}
\hline
 & PCam & CRC  & TILS & MHIST & MIDOG & \begin{tabular}[c]{@{}c@{}}OOD\\ Average\end{tabular} \\ \hline \hline
Standard DINOv2         & 83.2 & 92.1 &  91.7 & \textbf{80.1}  & 62.1  & 81.8       \\
\quad+ECT             & 82.3 {\color[HTML]{CB0000} $\downarrow$ 0.9} & 91.3 {\color[HTML]{CB0000} $\downarrow$ 0.8} & 91.1 {\color[HTML]{CB0000} $\downarrow$ 0.6} & 74.0  {\color[HTML]{CB0000} $\downarrow$ 6.1} & 63.3  {\color[HTML]{32CB00} $\uparrow$ 1.2} & 80.4  {\color[HTML]{CB0000} $\downarrow$ 1.4}     \\
\quad+KDE             & 84.2 {\color[HTML]{32CB00} $\uparrow$ 1.0} & 95.8 {\color[HTML]{32CB00} $\uparrow$ 3.7} &  \textbf{93.1} {\color[HTML]{32CB00} $\uparrow$ 1.4} & \textbf{81.3}  {\color[HTML]{32CB00} $\uparrow$ 1.3} & \textbf{66.4}  {\color[HTML]{32CB00} $\uparrow$ 4.2} & 84.2   {\color[HTML]{32CB00} $\uparrow$ 2.4}    \\
\quad-SOL             & 83.3 {\color[HTML]{32CB00} $\uparrow$ 0.1} & 92.9 {\color[HTML]{32CB00} $\uparrow$ 0.8} & 92.2 {\color[HTML]{32CB00} $\uparrow$ 0.5} & 77.5 {\color[HTML]{CB0000} $\downarrow$ 2.6} & 65.5  {\color[HTML]{32CB00} $\uparrow$ 3.3}& 82.3  {\color[HTML]{32CB00} $\uparrow$ 0.5}     \\
\quad+ECT,+KDE       & 86.6 {\color[HTML]{32CB00} $\uparrow$ 3.4} & 96.3 {\color[HTML]{32CB00} $\uparrow$ 4.3} & 92.9 {\color[HTML]{32CB00} $\uparrow$ 1.2} & \textbf{79.9} {\color[HTML]{CB0000} $\downarrow$ 0.2} & \textbf{67.4}  {\color[HTML]{32CB00} $\uparrow$ 5.2} & 84.6  {\color[HTML]{32CB00} $\uparrow$ 2.8}     \\
\quad+ECT,+KDE,-SOL & \textbf{87.3} {\color[HTML]{32CB00} $\uparrow$ 4.2} & \textbf{96.9} {\color[HTML]{32CB00} $\uparrow$ 4.9} &  \textbf{93.1} {\color[HTML]{32CB00} $\uparrow$ 1.4} & 78.8 {\color[HTML]{CB0000} $\downarrow$ 1.3} & 65.6 {\color[HTML]{32CB00} $\uparrow$ 3.5} & 84.4  {\color[HTML]{32CB00} $\uparrow$ 2.6} \\ \hline   
\end{tabular}
\caption{Weighted F1 score (multiplied by 100) for the out-of-domain tasks where the embedding is [CLS] only. ECT is the proposed extended context translation, KDE is the proposed regularization function and SOL refers to the removal of solarization.}
\label{tab:abalation_ood_cls_only}
\end{table}

\clearpage
\subsection{Scaling plots}
\label{sec:appendix_scaling}

\begin{figure}[h]
    \centering
    \includegraphics{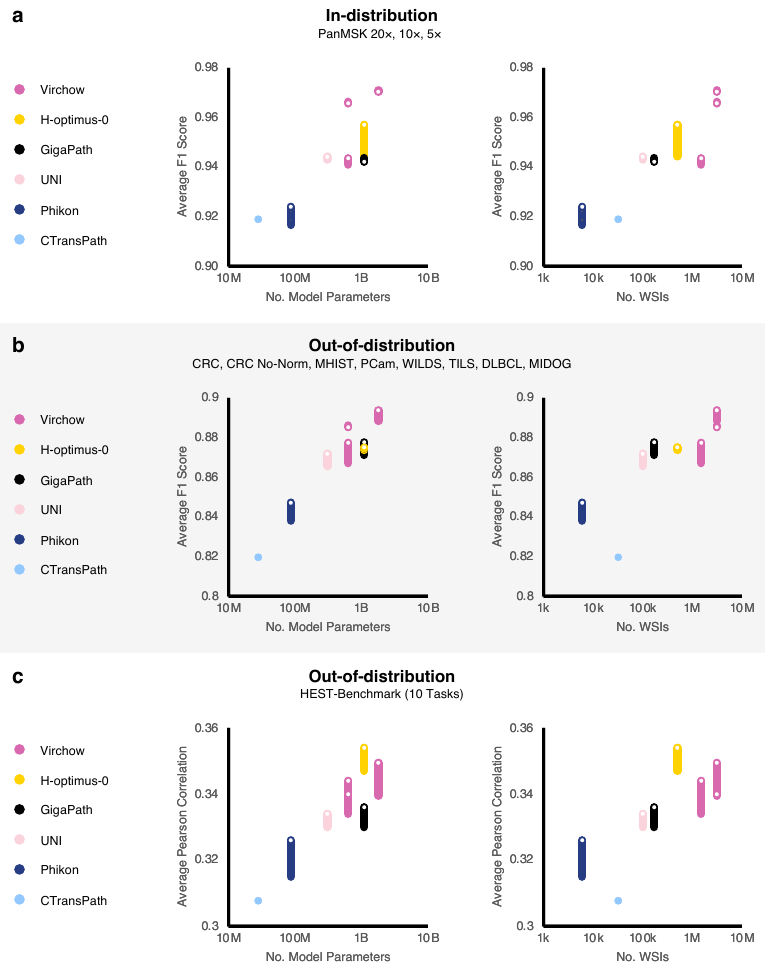}
    \caption{The average performance across in-distribution tasks (a) and out-of-distribution tasks (b, c) scales with model size (left) and training data size (right). In-distribution data is sourced from the same institutions as Virchow training data, without any patient-level overlap. For all models except CTransPath, which doesn't have a CLS token, we show the performance using CLS only (solid point) and the concatenated embedding (white dot). The connection between the two provides a visual representation of the change in performance. In most cases, using the concatenated representation improves performance.}
    \label{fig:all_scaling}
\end{figure}

\clearpage
\subsection{MIDOG patch token analysis}
\label{app_sec:patch-tokens}

The mitosis detection task (MIDOG) benefitted substantially from concatenating the mean of the patch tokens to the CLS embedding (CLS+Mean) in initial analysis. To further interrogate the value of the patch tokens, we trained models which aggregate the patch tokens to make a prediction. We test max-pooling and attention-based aggregation~\cite{ilse2018attention}. The results are in Table~\ref{tab:sup-patch}. In all cases, using a patch-based aggregator improves performance. This is likely because mitosis is a localized phenomenon. The larger models (Virchow2G and H-optimus-0) have the highest performance overall. These results suggest that patch tokens may contain additional information useful for some tasks and that some models may be better at summarizing this localized patch-level information in the CLS token than others; however, further study is necessary.

\begin{table}[h]
\centering 
{\renewcommand{\arraystretch}{1.15} 
\begin{tabular}{lcccc} \toprule
            & \begin{tabular}[c]{@{}c@{}}CLS-\\Only\end{tabular}  & \begin{tabular}[c]{@{}c@{}}CLS+\\Mean\end{tabular} & \begin{tabular}[c]{@{}c@{}}Patch Token\\ Max. Agg.\end{tabular} & \begin{tabular}[c]{@{}c@{}}Patch Token\\ ABMIL~\cite{ilse2018attention} \end{tabular} \\ \hline \hline
Virchow2G   & 80.5 & 83.6      & \textbf{87.5}                   & 86.7                 \\
Virchow2    & 80.0 & 80.4      & \textbf{86.4}                   & 85.6                 \\
Virchow     & 76.0 & 78.7      & \textbf{84.6}                   & 84.5                 \\
H-optimus-0 & 78.1 & 79.6      & \textbf{87.5}                   & 86.8                 \\
GigaPath    & 78.2 & 79.4      & \textbf{84.7}                   & 84.3                 \\
UNI         & 74.8 & 75.9      & 83.0                   & \textbf{83.8}                 \\
Phikon      & 76.9 & 73.0      & \textbf{80.6}                   & 80.3                 \\  \hline                
\end{tabular} }
\caption{Weighted F1 score results for different approaches for the MIDOG task using the ViT-based CPath FMs. Here we bold the best performing result for each FM. Note that results have been multiplied by 100 for improved readability.}
\label{tab:sup-patch}
\end{table}

\subsection{ViT-S distillation vs. from scratch}
\label{app_sec:vit-s}

The success of distilling Virchow2G into a ViT-S sized model (Virchow2G Mini) raises the question of whether distillation is necessary, or could the same recipe for Virchow2/2G allow for training a performant ViT-S from scratch without the need of a large pretrained teacher model? To investigate this, we trained a ViT-S/14 without distillation and compare to Virchow2G Mini after both models have seen 1B tiles. The results in Table~\ref{tab:vit-s-ablation-id} and Table~\ref{tab:vit-s-ablation-ood} show that distilling from Virchow2G leads to substantial performance gains across all tasks.

\begin{table*}[ht]
\centering
\small
{\renewcommand{\arraystretch}{1.15} 
\begin{tabular}{lllll}
\toprule
Model & PanMSK $20\times$ & PanMSK $10\times$ & PanMSK $5\times$ & Average \\ 
\hline
\hline
ViT-S scratch & 91.2 & 92.8 & 93.2 & 92.4 \\
Virchow2G Mini & \textbf{95.0} {\color[HTML]{32CB00} \tiny $\uparrow$ 3.8}  & \textbf{95.7} {\color[HTML]{32CB00} \tiny $\uparrow$ 2.9} & \textbf{96.0} {\color[HTML]{32CB00} \tiny $\uparrow$ 2.8} & \textbf{95.6} {\color[HTML]{32CB00} \tiny $\uparrow$ 3.2} \\
\bottomrule
\end{tabular}
}
\caption{Weighted F1 score (multiplied by 100) results for the in-distribution tile-level benchmark tasks.}
\label{tab:vit-s-ablation-id}
\end{table*}

\begin{table*}[ht]
\centering
\small
{\renewcommand{\arraystretch}{1.15} 
\begin{tabular}{llllllllll}
 \toprule
Model & CRC & {\begin{tabular}[c]{@{}c@{}}CRC\\ No-Norm\end{tabular}} & MHIST & PCam & WILDS & TILS & DLBCL & MIDOG & Average \\ 
\hline\hline
ViT-S scratch & 96.1 & 88.6 & 79.7 & 88.8 & 95.2 & 93.5 & 53.1 & 66.3 & 82.7 \\
Virchow2G Mini & \textbf{97.0} {\color[HTML]{32CB00} \tiny $\uparrow$ 0.9} & \textbf{96.0} {\color[HTML]{32CB00} \tiny $\uparrow$ 7.4} & \textbf{83.2} {\color[HTML]{32CB00} \tiny $\uparrow$ 3.5} & \textbf{93.5} {\color[HTML]{32CB00} \tiny $\uparrow$ 4.7} & \textbf{98.4} {\color[HTML]{32CB00} \tiny $\uparrow$ 3.2} & \textbf{94.1} {\color[HTML]{32CB00} \tiny $\uparrow$ 0.6} & \textbf{58.9} {\color[HTML]{32CB00} \tiny $\uparrow$ 5.8} & \textbf{79.6} {\color[HTML]{32CB00} \tiny $\uparrow$ 13.3} & \textbf{87.6} {\color[HTML]{32CB00} \tiny $\uparrow$ 4.9} \\ 
\bottomrule
\end{tabular}
}
\caption{Weighted F1 score (multiplied by 100) results for the out-of-distribution tile-level benchmark tasks.}
\label{tab:vit-s-ablation-ood}
\end{table*}

\end{document}